\documentclass[final,12pt,hidelinks]{elsarticle}

\usepackage{amssymb}
\usepackage{color}
\usepackage{multirow}
\usepackage{booktabs}
\usepackage{subfigure}
\usepackage{gensymb}
\usepackage{hyperref}
\usepackage{lineno}
\usepackage{epstopdf}

\journal{Deep Learning for Remotely Sensed Data}

\begin{document}

\begin{frontmatter}



\title{Algorithms for Semantic Segmentation of Multispectral Remote Sensing Imagery using Deep Learning}

\author{Ronald Kemker}
\author{Carl Salvaggio}
\author{Christopher Kanan}

\address{Chester F. Carlson Center for Imaging Science\\Rochester Institute of Technology\\54 Lomb Memorial Drive\\Rochester NY 14623\\
{\tt\small \{rmk6217, cnspci, kanan\}@rit.edu}}

\begin{abstract}
Deep convolutional neural networks (DCNNs) have been used to achieve state-of-the-art performance on many computer vision tasks (e.g., object recognition, object detection, semantic segmentation) thanks to a large repository of annotated image data.  Large labeled datasets for other sensor modalities, e.g., multispectral imagery (MSI), are not available due to the large cost and manpower required.  In this paper, we adapt state-of-the-art DCNN frameworks in computer vision for semantic segmentation for MSI imagery. To overcome label scarcity for MSI data, we  substitute real MSI for generated synthetic MSI in order to initialize a DCNN framework.  We evaluate our network initialization scheme on the new RIT-18 dataset that we present in this paper.  This dataset contains very-high resolution MSI collected by an unmanned aircraft system.  The models initialized with synthetic imagery were less prone to over-fitting and provide a state-of-the-art baseline for future work.

\end{abstract}


\end{frontmatter}


\section{Introduction}
Semantic segmentation algorithms assign a label to every pixel in an image. In remote sensing, semantic segmentation is often referred to as image classification, and semantic segmentation of non-RGB imagery has numerous applications, such as land-cover classification~\cite{grss2017data}, vegetation classification~\cite{laliberte2011multispectral}, and urban planning~\cite{rottensteiner2012isprs,volpi2015semantic}. Semantic segmentation has been heavily studied in both remote sensing and computer vision. In recent years, the performance of semantic segmentation algorithms for RGB scenes has rapidly increased due to deep convolutional neural networks (DCNNs). To use DCNNs for semantic segmentation, they are typically first trained on large image classification datasets that have over one million labeled training images. Then, these pre-trained networks are then adapted to the semantic segmentation task. This two-step procedure is necessary because DCNNs that process high-resolution color (RGB) images have millions of parameters, e.g., VGG-16 has 138 million parameters~\cite{SimonyanZ14a}. Semantic segmentation datasets in computer vision are too small to find good settings for the \emph{randomly initialized} DCNN parameters (weights), and over-fitting would likely occur without the use of pre-trained networks. For example, to evaluate a semantic segmentation method on the RGB PASCAL VOC datasets~\cite{everingham2010pascal}, state-of-the-art methods use a DCNN pre-trained on ImageNet (1.28 million training images), fine-tune it for semantic segmentation on the COCO dataset (80K training images)~\cite{lin2014microsoft}, and then fine-tune it again on PASCAL VOC (1,464 training images)~\cite{ChenPK0Y16,LinMS016}.

\begin{figure}[t]
    \centering
    \includegraphics[width=0.7\linewidth]{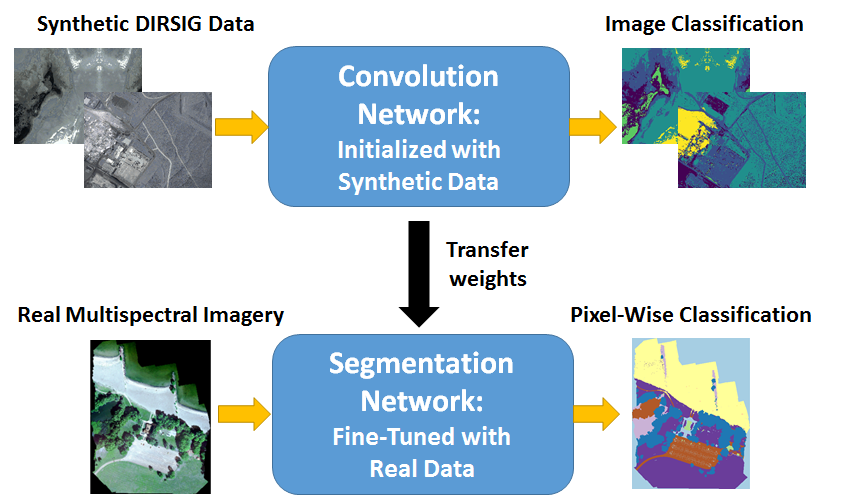}
	\label{fig:upfront}
    \caption{Our proposed model uses synthetic multispectral imagery to initialize a DCNN for semantic segmentation.  This model is then fine-tuned on real imagery.}
\end{figure}

Utilizing pre-trained networks to prevent overfitting works well for RGB imagery because massive labeled datasets are available; but in the non-RGB domain, label scarcity is a far greater problem. For example, existing semantic segmentation benchmarks for hyperspectral imagery consist of a single image mosaic. Therefore, pre-training DCNNs on hand-labeled datasets consisting of real images is not currently possible in non-RGB domains. In this paper, we explore an alternative approach: using vast quantities of automatically-labeled \emph{synthetic} multispectral imagery (MSI) for pre-training DCNN-based systems for semantic segmentation.  

\begin{figure}[t]
  \centering
  \subfigure[Train]{%
  \includegraphics[width=0.25\linewidth]{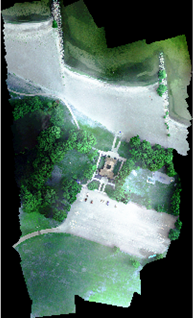}
  \label{fig:train_ortho}}
  \subfigure[Validation]{%
  \includegraphics[width=0.3225\linewidth]{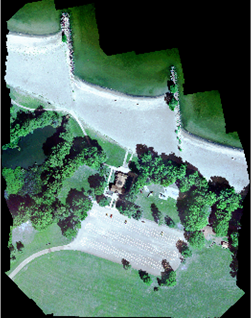}
  \label{fig:val_ortho}}
  \subfigure[Test]{%
  \includegraphics[width=0.253\linewidth]{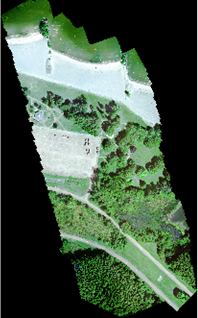}
  \label{fig:test_ortho}}
  \caption{RGB visualization of RIT-18 dataset.  This dataset has six spectral bands.}
  \label{fig:ortho}
\end{figure}

We propose to use the Digital Imaging and Remote Sensing Image Generation (DIRSIG) modeling software to generate large quantities of synthetic MSI and corresponding label maps.  We use DIRSIG to build a large, diverse scene model, in which we can simulate various weather and lighting conditions. We then capture synthetic aerial images of the scene with a MSI sensor model.  We use the synthetic data to initialize a DCNN for object recognition, and then we combine the pre-trained DCNN with two different fully-convolutional semantic segmentation models using real MSI. 

In the past, researchers have used DCNNs pre-trained on ImageNet to yield state-of-the-art results for the semantic segmentation of high-resolution multispectral aerial imagery~\cite{rs9050446, marmanis2016semantic} because the most widely used benchmarks~\cite{rottensteiner2012isprs} only use a single non-RGB band.  What happens when the spectral range of the dataset increases?  The real MSI used to evaluate our network initialization scheme comes from a new semantic segmentation dataset that we built called RIT-18\footnote{The dataset is available at \url{https://github.com/rmkemker/RIT-18}}.  RIT-18 consists of high-resolution MSI (six bands) acquired by an unmanned aircraft system (UAS). The primary use of this dataset is for evaluating semantic segmentation frameworks designed for non-RGB remote sensing imagery.  The dataset, shown in Fig. \ref{fig:ortho}, is split into training, validation, and testing folds to 1) provide a standard for state-of-the-art comparison, and 2) demonstrate the feasibility of deploying algorithms in a more realistic setting.  Baseline results demonstrate that the large spatial variability commonly associated with high-resolution imagery, large sample (pixel) size, small and hidden objects, and unbalanced class distribution make this a difficult dataset to perform well on, making it an excellent dataset for evaluating our DCNN frameworks for semantic segmentation.

\textbf{Contributions}: Our paper makes three major contributions:
1) We are the first to adapt recent fully-convolutional DCNNs to semantic segmentation of multispectral remote sensing imagery; 2) We demonstrate that pre-training these networks on synthetic imagery can significantly improve their performance; and 3) We describe the new RIT-18 dataset for evaluating MSI semantic segmentation algorithms.

\section{Related Work}

\subsection{Semantic Segmentation of RGB Imagery with Deep Networks}
In this paper, “pixel-wise classification” and “semantic segmentation” are synonymous.  Semantic segmentation is the term more commonly used in computer vision and is becoming increasingly used in remote sensing.  State-of-the-art semantic segmentation frameworks for RGB imagery are trained end-to-end and consist of convolution and segmentation sub-networks.  The convolution network is usually a pre-trained DCNN designed to classify images from ImageNet ~\cite{long2015fully,NohHH15,ChenPK0Y16,PinheiroLCD16}, and current state-of-the-art performers use VGG-16~\cite{SimonyanZ14a} or ResNet~\cite{ILSVRC15}.  The segmentation network is appended to the convolution network and is designed to reconstruct the feature response to the same spatial dimensions as the input before assigning semantic labels.  The resulting semantic segmentation network can be fine-tuned with orders of magnitude fewer training images (thousands versus millions of images) because the convolutional network is already trained. We describe some of the best performing recent models below, all of which used non-aerial RGB scenes.

The first fully-convolutional network (FCN) designed for semantic segmentation~\cite{long2015fully} used the VGG-16 network~\cite{SimonyanZ14a}, which has approximately 138 million parameters.   VGG-16 is trained to do image classification on ImageNet~\cite{ILSVRC15}, rather than directly for semantic segmentation. Their FCN model used coarse upsampling and deconvolution in the segmentation network to classify each pixel.  The net's major disadvantage was that VGG-16's 5 max-pooling layers shrunk the original image by a factor of 32, resulting in a coarse label map~\cite{long2015fully}.

In \cite{NohHH15}, they proposed to improve the FCN model by building a symmetric (deconvolution) network using spatial unpooling and deconvolution layers.  This increased performance when classifying objects at multiple resolutions (i.e. small or large objects in the image); however, it still produced a coarse label map.  As a post-processing step, the authors used a conditional random field (CRF) to sharpen the classification boundaries~\cite{koltun12}. The major downside to this deconvolution network was that it required more memory and time to train compared to \cite{long2015fully}. 

The DeepLab semantic segmentation network~\cite{ChenPK0Y16}  was built with the ResNet DCNN. DeepLab mitigates downsampling issues and makes segmentation boundaries sharper by replacing conventional convolution layers with atrous convolutions.  An atrous convolution filter is filled with zeros between the sample points; so although the effective size of the filter increases, the number of trainable parameters remains constant.  When these filters are convolved with an image, it can preserve the original dimensions. The authors found that using atrous filters throughout the entire network was inefficient, so they used both conventional and atrous filters, reducing the image only by a factor of eight. A dense CRF was used as a post-processing step to make the predicted label map sharper. 

Many of these earlier models used a CRF as a post-processing step to sharpen the classification masks, but it may be better to allow the network to directly optimize itself towards creating a sharper label mask. Two recent models that did this, Sharpmask~\cite{PinheiroLCD16} and RefineNet~\cite{LinMS016}, used skip-connections to incorporate image refinement into the end-to-end model.  Sharpmask used a refinement module to combine features from the convolution network with the upsampled features from the segmentation network.  RefineNet improved the Sharpmask model with multi-resolution fusion (MRF) to combine features at different scales, chained residual pooling (CRP) to capture background context, and residual convolutional units (RCUs) to improve end-to-end training.  In this paper, we adapt the Sharpmask and RefineNet models to multispectral remote sensing imagery and use them to evaluate our proposed initialization procedure.  These DCNN algorithms are described in more detail in Sections \ref{section:sharpmask} and \ref{section:refinenet}. 

\subsection{Deep-Learning for Non-RGB Sensors}
Deep learning approaches to classify and analyze RGB remote sensing imagery have advanced considerably thanks to deep learning, including the use of DCNNs pre-trained on ImageNet and unsupervised feature extraction~\cite{cheng2017remote,cheng2016learning,yao2016semantic}.  Deep-learning frameworks for the semantic segmentation of multispectral and hyperspectral images have been explored by the remote sensing community; however, the paucity of annotated data available for these sensor modalities has pushed researchers to embrace unsupervised feature extraction methods~\cite{Kemker16} and object based image analysis~\cite{chen2017geographic} (see Section \ref{section:geobia}).  Deep features extracted from every pixel of the labeled data are then used with a classifier, often a support vector machine, to generate a pixel-wise classification map.  

The authors in \cite{Khatami201689} determined that spatial (texture) information influenced classification performance the most, which is why current state-of-the-art methods extract spatial-spectral features from the image data.  Early spatial-spectral feature extractors had hyperparameters that required tuning (e.g. gray-level co-occurrence matrices~\cite{mirzapour2015improving}, Gabor~\cite{shen2013discriminative}, sparse coding~\cite{soltani2015pixels}, extended morphological attribute profiles~\cite{ghamisi2014automatic}, etc).  These hand-crafted features could fail to generalize well across multiple datasets, so they were replaced with learned features that were automatically tuned from the data itself.

Arguably, the most successful of these learned feature extraction methods for remote sensing imagery is the stacked autoencoder~\cite{liu2015hyperspectral,lin2013spectral,zhao2015combining,ma2015hyperspectral,tao2015unsupervised}.  An autoencoder is an unsupervised neural network that learns an efficient encoding of the training data. These autoencoders can be stacked together to learn higher feature representations.

In \cite{tao2015unsupervised}, stacked sparse autoencoders (SSAE) were used to learn feature extracting filters from one hyperspectral image, and then these filters were used to classify the same image as well as a different target image.  The SSAE features learned on the one dataset failed to properly transfer to another dataset because these features failed to generalize well.  One possible way to overcome this limitation would be to do the unsupervised training on multiple images, which may lead to more discriminative features that generalize between different target datasets.

This idea was explored in \cite{Kemker16}, where the authors showed that training on large quantities of unlabeled hyperspectral data, which is known as self-taught learning~\cite{raina2007self}, improved classification performance. The authors proposed two different frameworks for semantic segmentation that used self-taught learning: multi-scale independent component analysis (MICA) and stacked convolutional autoencoders (SCAE). SCAE outperformed MICA in their experiments, but both approaches advanced the state-of-the-art across three benchmark datasets, showing that self-taught features can work well for multiple images.

\subsection{Semantic Segmentation of High Resolution Remote Sensing Imagery}
\label{section:geobia}
As the resolution of available image data increased, researchers explored geographic object based image analysis (GEOBIA) to deal with the high spatial variability in the image data~\cite{blaschke2010object}.  According to \cite{hay2008geobia}, GEOBIA involves the development of ``automated methods to partition remote sensing imagery into meaningful image-objects, and assessing their characteristics through spatial, spectral and temporal scales, so as to generate new geographic information in GIS-ready format.''  The model can generate superpixels through clustering (unsupervised) or segmenting (supervised) techniques, extrapolate information (features) about each superpixel, and use that information to assign the appropriate label to each superpixel.  Clustering/Segmenting the image into superpixels could prevent the salt-and-pepper misclassification errors that are characteristic of high-resolution imagery.  Another strategy for mitigating this type of error is post-processing the classification map with a Markov Random Field~\cite{gewali2016spectral} or CRF~\cite{zhong2007multiple}.

Classifying superpixels is another semantic segmentation strategy found in the computer vision literature~\cite{farabet2013learning,hariharan2014simultaneous}, but it is less used with the development of fully-convolutional neural networks for end-to-end semantic segmentation~\cite{long2015fully}.  In contrast, remote sensing researchers have been very successful employing this strategy, via GEOBIA, for segmenting high resolution imagery.  Our DCNN approach had multiple advantages: 1) it did not require a time-consuming unsupervised segmentation, 2) it could be trained end-to-end, 3) it increased classification performance, and 4) it can be done much faster. Incorporating a CRF or employing GEOBIA methods could tighten classification map boundaries and reduce salt-and-pepper label noise~\cite{koltun12}.  In addition, skip-connections that pass low-level features to the segmentation network were shown to tighten classification boundaries and reduce salt-and-pepper errors in end-to-end semantic segmentation frameworks~\cite{PinheiroLCD16,LinMS016}.  In our paper, we chose to use end-to-end, fully-convolutional networks with skip-connections for classification, but future work could include a CRF and/or GEOBIA methods to further increase performance.
\subsection{MSI Semantic Segmentation Datasets for Remote Sensing}

Most remote sensing semantic segmentation datasets have been imaged by airborne or satellite platforms. The existing publicly available MSI datasets are shown in Table~\ref{table:msi}. The gold-standard benchmark for the semantic segmentation of visual near-infrared (VNIR) MSI are the Vaihingen and Potsdam datasets hosted by the International Society for Photogrammetry and Remote Sensing (ISPRS)~\cite{rottensteiner2012isprs}.  These datasets have comparably high spatial resolution, but only five classes.  Newer datasets, such as Zurich~\cite{volpi2015semantic} and EvLab-SS~\cite{zhang2017learning}, have sacrificed some spatial resolution but include additional labeled classes.  Additional competitions hosted by the IEEE Geoscience and Remote Sensing Society (GRSS)~\cite{grss2017data} and Kaggle~\cite{kaggle2017} involve the fusion of multi-modal imagery captured (or resampled) to different ground sample distances (GSDs). 

In this paper, we describe a new semantic segmentation dataset named RIT-18.  Some of the advantages of our dataset over existing benchmarks include:

\begin{enumerate}
\item RIT-18 is built from \textbf{very-high resolution} (4.7 cm GSD) UAS data.  The ISPRS datasets are the only comparable datasets, but these datasets only use 3-4 spectral bands and 5 object classes.
\item RIT-18 has \textbf{18 labeled object classes}.  The 2017 IEEE GRSS Data Fusion has 17 object classes, but all of the data has been spatially resampled to 100 meter GSD (land-cover classification).
\item RIT-18 has \textbf{6 VNIR spectral bands}, including two additional NIR bands not included in the ISPRS benchmarks.  These additional bands increase the discriminative power of classification models in vegetation heavy scenes (see Table \ref{table:band_selection}.)
\item RIT-18 was \textbf{collected by a UAS platform}.  A UAS is cheaper and easier to fly than manned aircraft which can allow the end-user to collect high spatial resolution imagery more frequently.
\item RIT-18 is a \textbf{very difficult dataset} to perform well on.  It has an unbalanced class distribution.  A system that is capable of performing well on RIT-18 must be capable of low-shot learning.
\end{enumerate}

\begin{table}[ht!]
\centering \footnotesize
\begin{tabular}{@{}lcccc@{}}\toprule
\textbf{Dataset} & \textbf{Year} & \textbf{Sensor(s)} &\textbf{GSD} & \textbf{Classes}  \\ \midrule \small
ISPRS Vaihingen~\cite{rottensteiner2012isprs}&2012 & Green/Red/IR  & 0.09 & 5  \\
ISPRS Potsdam~\cite{rottensteiner2012isprs}&2012 & 4-band (VNIR)  & 0.05 & 5  \\
Zurich Summer~\cite{volpi2015semantic}&2015 & QuickBird & 0.61 & 8 \\
\multirow{ 2}{*}{EvLab-SS~\cite{zhang2017learning}}&\multirow{ 2}{*}{2017} & World-View-2, GF-2,  & \multirow{ 2}{*}{0.1-1.0} & \multirow{ 2}{*}{10} \\
& &QuickBird, \& GeoEye& & \\
\multirow{ 2}{*}{GRSS Data Fusion~\cite{grss2017data}}&\multirow{ 2}{*}{2017} & Landsat \& &\multirow{ 2}{*}{100} & \multirow{ 2}{*}{17}  \\
& &Sentinel 2& & \\
Kaggle Challenge~\cite{kaggle2017}& 2017 & World-View 3 & 0.3-7.5 & 10 \\
\textbf{RIT-18}& \textbf{2017} & \textbf{6-band (VNIR)} & \textbf{0.047} & \textbf{18} \\
\bottomrule \\
\end{tabular}
\caption{Benchmark MSI semantic segmentation datasets, the year they were released, the sensor that collected it, its ground sample distance (GSD) in meters, and the number of labeled object classes (excluding the background class).  Our RIT-18 dataset is in bold.}
\label{table:msi}
\end{table}

\subsection{Deep Learning with Synthetic Imagery}

Synthetic data has been used to increase the amount of training data for systems that use deep neural networks, and this has been done for many applications, including object detection~\cite{peng2015learning}, pose estimation~\cite{chen2016synthesizing}, face and hand-writing recognition~\cite{zhang2015learning}, and semantic segmentation~\cite{ros2016synthia}. Past work used various methods to generate large quantities of synthetic data including geometric/color transformations, 3D modeling, and virtual reality emulators.

The major upside to synthetic imagery is that it is normally cheaper and easier to obtain than images that are manually annotated by humans; however, the difference in feature-space distributions, also known as the synthetic gap, can make it difficult to transfer features from synthetic to real imagery.  Researchers have adopted domain adaptation techniques~\cite{ben2010theory} to mitigate this phenomenon, including training autoencoders to shift the distribution of the synthetic data to the distribution of the real data~\cite{guyon2011unsupervised, zhang2015learning}.  This can allow for a classifier to be trained using synthetic images and then make predictions on real data.

Network fine-tuning, has been widely-adopted in the computer vision community for re-purposing DCNNs trained for image classification for a variety of alternative applications, such as semantic segmentation.  Typically, a portion of the network is pre-trained on a large image classification dataset (e.g., ImageNet), and then adapted to a dataset with different class labels and feature distributions. However, it is possible to use synthetic data instead. In \cite{ros2016synthia}, the authors built a synthetic dataset using a virtual reality generator for the semantic segmentation of autonomous driving datasets, and then they combined the synthetic and real imagery to train their semantic segmentation model.  In this paper, we initialized the ResNet-50 DCNN to classify synthetic MSI, and then we fine-tuned these weights to perform semantic segmentation on real MSI. 

\section{Methods}

In this section, we first describe how we generated synthetic imagery for pre-training DCNNs to classify MSI data. Then we describe two fully convolutional semantic segmentation algorithms, which we adapt for MSI imagery. Lastly, we describe both simple baseline and state-of-the-art algorithms for semantic segmentation for remote sensing imagery, which will serve as a comparison to the FCN models.

\subsection{Synthetic Image Generation using DIRSIG}

Because there are no publicly-available ImageNet sized datasets for non-RGB sensor modalities, we used DIRSIG to build a large synthetic labeled dataset for the semantic segmentation of aerial scenes.  DIRSIG is a software tool used heavily in the remote sensing industry to model imaging sensors prior to development.  It can be used to simulate imaging platforms and sensor designs, including monochromatic, RGB, multispectral, hyperspectral, thermal, and light detection and ranging (LIDAR). 

DIRSIG images an object/scene using physics-based radiative transfer/ propagation modeling~\cite{Ientilucci03,schott1999advanced}.  A realistic object can be defined in the DIRSIG environment with 3D geometry, textures, bi-directional reflection distribution function (BRDF), surface temperature predictions, etc.  A custom sensor (e.g. MSI sensor) and platform (e.g. UAS) can be defined in the DIRSIG environment to ``fly-over'' a scene and image it.  The position of the sun/moon, the atmosphere profile, and the flight plan can all be modified to generate realistic data as well as provide a pixel-wise classification map. 

\begin{figure}[t]
\begin{center}
   \includegraphics[width=0.6\linewidth]{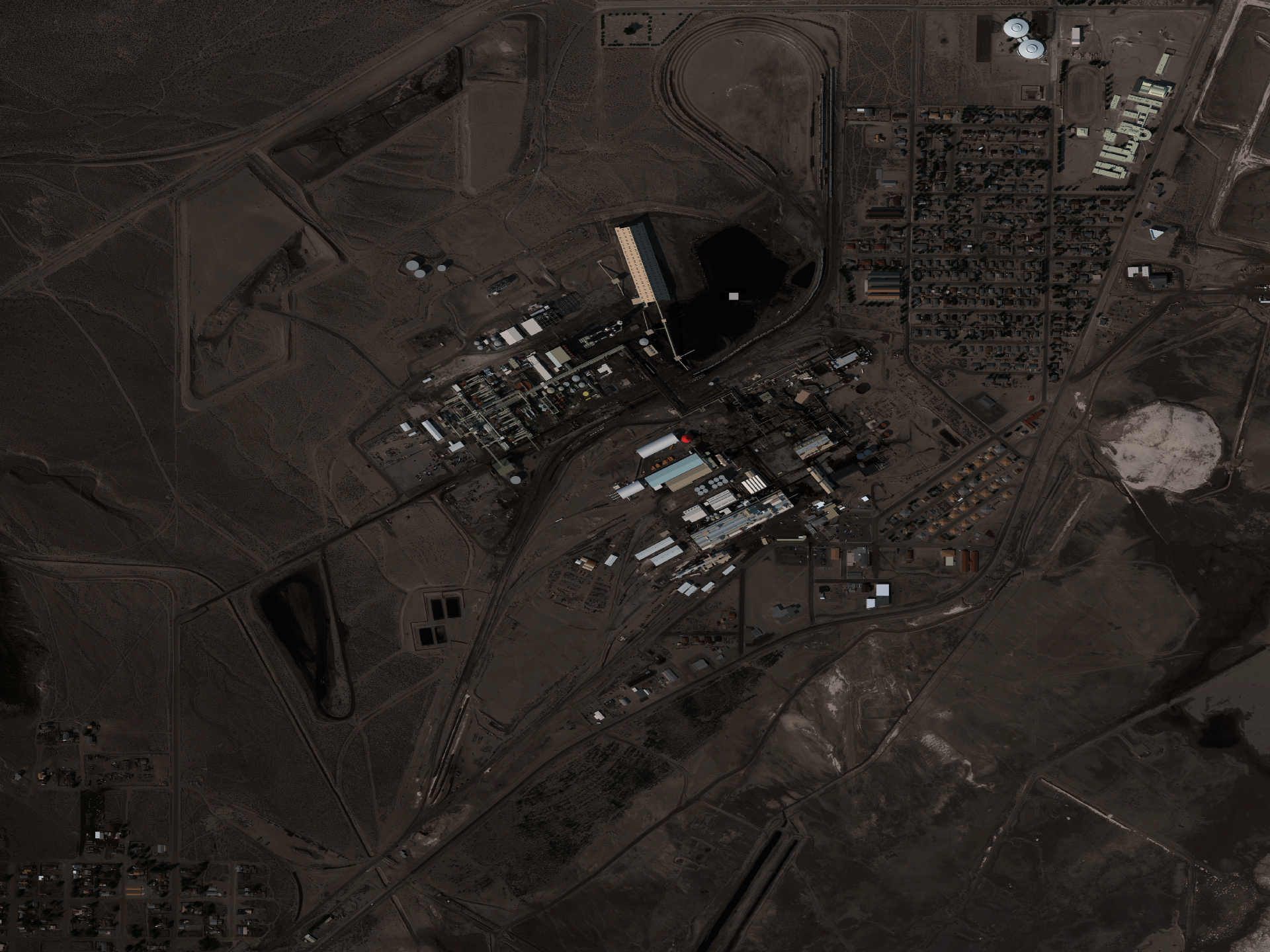}
\end{center}
   \caption{The DIRSIG Trona Scene.}
\label{fig:trona}
\end{figure}

We used the synthetic scene shown in Fig.~\ref{fig:trona}, which resembles Trona, an unincorporated area in Southern California. It is an industrial and residential scene containing 109 labeled classes, including buildings, vehicles, swimming pools, terrain, and desert plant life.

Trona is only accurate to a GSD of 0.5 meters, and creating images with smaller GSDs produces image artifacts.  The GSD of RIT-18 is about ten times higher; however, Trona was the only available scene that was 1) sufficiently large, 2) had enough object classes, and 3) had an accurate ground truth map.  To mitigate this problem, we forced our networks to learn scale-invariant features by generating MSI at multiple GSDs (0.5, 0.75, and 1 meter).  This is accomplished by flying the simulated UAS at different elevations.

We ``flew'' the drone across the entire synthetic scene with some overlap to make sure all objects located at the edges of some images are located near the center of other images. We varied the time-of-year (summer and winter), time-of-day (morning and afternoon), and the corresponding atmospheric conditions (mid-latitude summer and winter).  The semantic label for each pixel is the dominant class that lies within the instantaneous field of view of that pixel.  We built the final synthetic dataset by breaking the scene into smaller image patches, resulting in 4.7 million training and 520 thousand validation 80x80 MSI.  The label for each image is the majority category present.

\subsection{Fully-Convolutional Deep Networks for Semantic Segmentation}

There are several ways to sharpen the object boundaries in a classification map.  Post-processing techniques such as a fully-connected CRF can be used to eliminate small errors and sharpen object boundaries.  GEOBIA methods use the superpixel boundaries to help sharpen the output mask.  The authors in \cite{zhao2017object} combined CNN features with superpixel features to increase the detail in the object boundaries and reduce salt-and-pepper noise.  Both CRFs and GEOBIA methods require an additional step to sharpen the output mask which adds additional processing time; and in many cases, requires tuning of additional hyperparameters.  In this paper, we adapted two recent fully-convolutional deep neural networks for the semantic segmentation of MSI: SharpMask~\cite{PinheiroLCD16} and RefineNet~\cite{LinMS016}.  Both of them produced state-of-the-art results on standard semantic segmentation benchmarks in computer vision.  The Sharpmask and RefineNet models were designed to learn the mask sharpening operation in their respective end-to-end frameworks without post-processing with a CRF or clustering/segmenting the image.  This is done by passing lower level features from the DCNN to the parts of the segmentation network that are responsible for upsampling the feature map. This restores high spatial frequency information (such as object boundaries or other detail) that was lost when the feature map was downsampled.

The DCNN used for both of these models was ResNet-50~\cite{HeZRS15} with the improved network architecture scheme proposed in \cite{HeZR016}, where batch-normalization and ReLU are applied prior to each convolution. Both models were implemented using Theano/Keras~\cite{chollet2015keras} and trained on computers with a NVIDIA GeForce GTX Titan X (Maxwell) graphical processing unit (GPU), Intel Core i7 processor, and 64 GB of RAM.  Our goal is to compare the performance of these algorithms to baseline and state-of-the-art methods for semantic segmentation of remote sensing imagery, and to measure the benefit of pre-training with synthetic imagery.

We describe the architectural details of these two models in this section, but details regarding training of these networks is given in Section~\ref{section:net-training}.

\subsubsection{SharpMask}
\label{section:sharpmask}

\begin{figure}[t]
\begin{center}
   \includegraphics[width=0.6\linewidth]{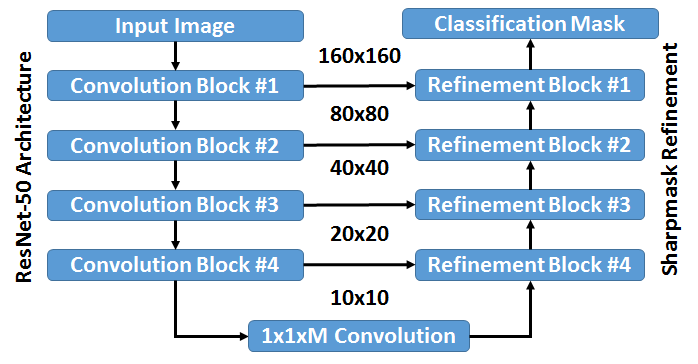}
\end{center}
   \caption{Our Sharpmask model. The spatial dimensions of each module's output are provided to illustrate that the refinement layer combines features from the feed-forward network (convolutional blocks) and the upsampled features in the reverse network to restore the original image dimensions.  The output is a class probability matrix for each pixel.} 
\label{fig:sharpmask}
\end{figure}

The Sharpmask model~\cite{PinheiroLCD16} used for this paper is illustrated in Fig.~\ref{fig:sharpmask}.  The network is broken into the convolution, bridge, and segmentation sub-networks.   Note that Sharpmask does not use all of the ResNet-50 model. As shown in Fig.~\ref{fig:sharpmask}, it only uses the first four macro-layers.  A macro-layer contains the convolution, batch-normalization, and ReLU activation layers right up to where the feature map is down-sampled by a factor of 2x.  This corresponds to the first 40 ResNet-50 convolution layers.  SharpMask was developed by Facebook to be lightweight and fast, possibly for deployment on a platform where size, weight, and power (SWaP) constraints are limited (e.g. UAS, embedded platform).  They tested and compared the classification performance and prediction time for models with various capacities including models that used only the first three and four ResNet macro-layers.  This trade-off study showed that SharpMask with four macro-layers provided the desired classification accuracy while also providing a 3x speed-up to their previous state-of-the-art DeepMask model\cite{pinheiro2015learning}.  We slightly modified the SharpMask model to retain the batch normalization layers for regularization.

The bridge network is a $M \times 1 \times 1$ convolution layer between the convolution and segmentation networks, where $M$ is selected as a trade-off between performance and speed.  The main goal of this network is to add some variability to the features fed into the segmentation network at refinement module \#4.  We use a value of $M=512$, which worked well in preliminary experiments. 

\begin{figure}[t]
\begin{center}
   \includegraphics[width=0.5\linewidth]{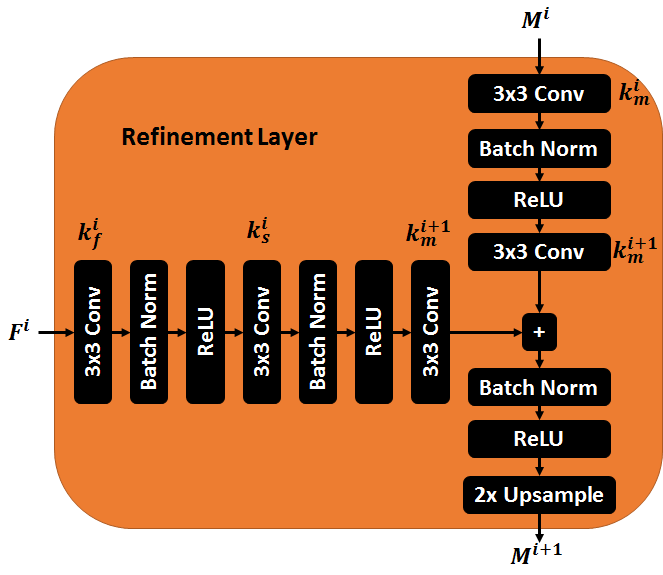}
\end{center}
   \caption{Sharpmask refinement with batch normalization. This layer learns mask refinement by merging features from the convolution $F^i$ and segmentation $M^i$ networks.}
\label{fig:refinement}
\end{figure}

The segmentation network uses refinement modules to restore the bridge layer output to the original dimensionality of the input data.  Instead of using a fully-connected CRF, the segmentation sharpening was learned as a part of the end-to-end network.  The refinement module merges low-level spatial features $F_i$ from the convolution network with high-level semantic content in the segmentation network $M_i$, as illustrated in Fig.~\ref{fig:refinement}.   

Each refinement module uses convolution and sum-wise merge layers prior to upsampling the feature response.  The upsampled feature response is fed into the refinement module at the next higher dimension.  The number of filters used in the $i$-th refinement module are given by
\begin{equation}
\label{eq:refinement}
k_s^i = k_m^i = \frac{128}{2^{i-1}} , i \leq 4.
\end{equation}
These parameters differ slightly from \cite{PinheiroLCD16} because the higher dimensionality of RIT-18 required a slightly larger model capacity. 

\subsubsection{RefineNet}
\label{section:refinenet}

\begin{figure}[t]
\begin{center}
   \includegraphics[width=0.6\linewidth]{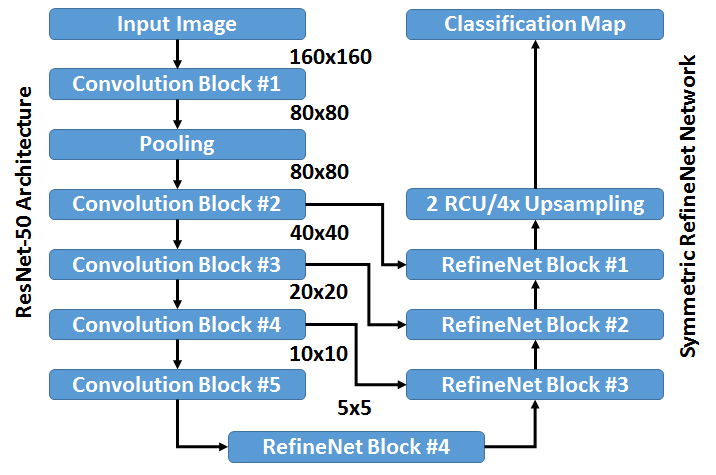}
\end{center}
   \caption{Network Architecture incorporating RefineNet modules for semantic segmentation.}
\label{fig:refinenet}
\end{figure}

\begin{figure}[t]
\begin{center}
  \subfigure[RefineNet Block]{%
  \includegraphics[width=0.55\linewidth]{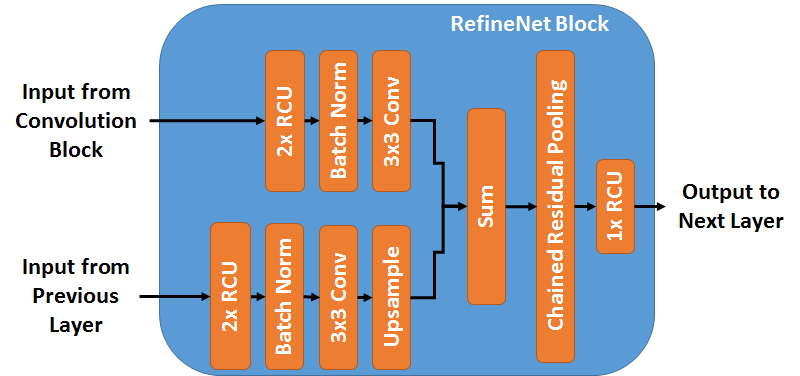}
  \label{fig:refinenet_block}}
  \subfigure[Residual Convolution Unit]{%
  \includegraphics[width=0.45\linewidth]{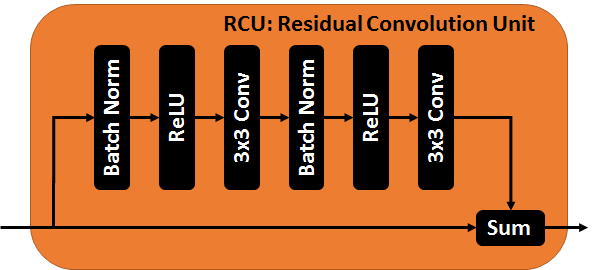}
  \label{fig:rcu}}
    \subfigure[Chained Residual Pooling with two window sizes]{%
  \includegraphics[width=0.45\linewidth]{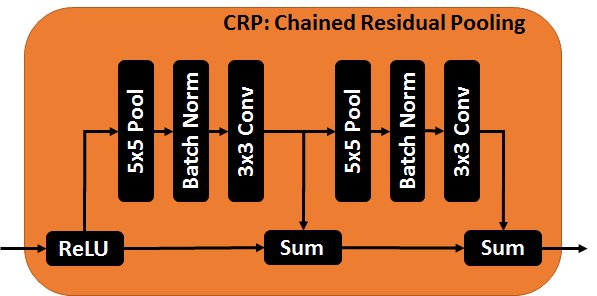}
  \label{fig:crp}}
  \end{center}
   \caption{RefineNet architecture.}
\label{fig:refinenetblock}
\end{figure}

The RefineNet model \cite{LinMS016} used in this paper (Fig.~\ref{fig:refinenet}) follows the same basic structure as the Sharpmask model with a few minor changes.  The refinement block in Fig.~\ref{fig:refinement} is replaced with a more complex block called RefineNet (Fig.~\ref{fig:refinenet_block}), which is broken up into three main components: residual convolution units (RCUs), multi-resolution fusion (MRF), and chained residual pooling (CRP). Our model uses batch normalization for regularization.  The convolutional network uses all five ResNet-50 macro-layers, so it is using every convolution layer in ResNet-50 except for the softmax classifier.

The RCUs (Fig.~\ref{fig:rcu}) are used to propagate the gradient across short- and long-range connections, which makes end-to-end training more effective and efficient.  The MRF is used to combine features at multiple scales, which in this paper, will be two.  The CRP module (Fig.~\ref{fig:crp}) pools features across multiple window sizes to capture background context, which is important for discriminating classes that are spatially and spectrally similar.  Fig.~\ref{fig:crp} uses two window sizes to illustrate how CRP works; however, our model pools features across four window sizes.

\subsection{Comparison Semantic Segmentation Algorithms}

Because RIT-18 is a new dataset, we compared the two FCN models to a number of classic  approaches, including classifying individual or mean-pooled pixels.  We also used two spatial-spectral feature extraction methods, MICA and SCAE, which recently achieved state-of-the-art performance on the semantic segmentation of hyperspectral imagery.   These methods use unsupervised learning to acquire spatial-spectral feature representations making them sample efficient.  As another baseline, we also used a method inspired by GEOBIA. 

\subsubsection{Simple Classification Algorithms}

One simple approach to semantic segmentation that has been widely used in remote sensing is running a classifier directly on each individual pixel. Using this approach, we establish baseline results using three different classifiers:  $k$-nearest neighbor (kNN), linear support vector machine (SVM), and multi-layer perceptron (MLP).  We also used spatial mean-pooling (MP) as a simple way of incorporating neighboring spatial information into the classifier.

The kNN classifier used the Euclidean distance metric and we cross-validated for $k$ over the range of 1-15.  

For the linear SVM, we used the LIBLINEAR implementation~\cite{fan2008liblinear}, which works well with large datasets. During training, it uses L2 regularization and adopts the one-vs-rest paradigm to multi-class classification. The input was scaled to zero-mean/unit-variance, where the mean and standard deviation were computed using the training data.  The SVM used RIT-18's validation set to cross-validate for its cost parameter $C$ over the range of $2^{-9} - 2^{16}$, and then the training and validation sets were combined to fit the final model.  The loss function was weighted by the inverse class frequency.

Our MLP implementation is a fully-connected neural network with a single hidden-layer of 64 units, chosen through cross-validation. This hidden layer is preceded by a batch-normalization layer and followed by a ReLU activation. To compensate for class unbalance, we assign each class distinct weights, which are given for class $i$ by
\begin{equation}
\label{eq:sample_weights}
w_{i} = \mu * \log_{10}{ \frac{\sum_{1}^N h_i}{h_i}},
\end{equation}
where $h_i$ is the number of pixels labeled as class $i$, $N$ is the number of classes ($N = 18$), and $\mu$ is a tunable parameter.  The MLP was trained using the NAdam optimizer \cite{KingmaB14}, with a batch size of 256, L2 regularization value of $10^{-4}$ in the convolution and batch normalization layers, and the class weighted update in Equation \ref{eq:sample_weights} where $\mu=0.15$.

Running classifiers on individual pixels ignores neighboring pixel information. This will negatively impact performance due to the high spatial variability commonplace in high resolution imagery. To address this, we also test the MP model.  We convolve the original data with a $5\times 5$ mean-pooling filter and then pass the response to the same Linear SVM described earlier.

\subsubsection{MICA}
\label{section:mica}
MICA achieved excellent performance at semantic segmentation of HSI data~\cite{Kemker16}, and we use it here as one of our baseline algorithms. MICA uses the unsupervised learning algorithm independent component analysis (ICA) to learn a spatial-spectral filter bank from images captured by the sensor. The filters that it acquires exhibit color opponency and resemble Gabor-type (bar/edge) filters. These filters are then convolved with the image, like a single layer convolutional neural network, and their responses are normalized using a non-linear activation function. These responses are then pooled to incorporate translation invariance, and then a classifier is applied to each of the pooled responses. While the original paper used an RBF-SVM to classify these responses, that is not feasible due to the size of RIT-18. Instead, we fed these responses into an MLP classifier. 

The MICA model used in this paper had $F=64$ learned filters of size $25 \times 25 \times 6$ and a mean pooling window of 13. The MLP model used to classify the final MICA feature responses used one hidden layer with 256 ReLU units and a softmax output layer.  The MICA filters were trained with 30,000 $25 \times 25$ image patches randomly sampled from the training and validation folds.  Additional details for the MICA algorithm can be found in \cite{Kemker16}.

\subsubsection{SCAE}
\label{section:scae_exp}
SCAE is another unsupervised spatial-spectral feature extractor that achieved excellent results on the semantic segmentation of hyperspectral imagery \cite{Kemker16}. SCAE extracts features using stacked convolutional autoencoders, which are pre-trained using unsupervised learning. SCAE has a deeper neural network architecture than MICA and is capable of extracting higher-level semantic information.  The SCAE model used in this paper is almost identical to the model proposed in \cite{Kemker16} with a few notable exceptions.  First, SCAE was trained with 30,000 $128\times 128$ image patches randomly extracted from the training and validation datasets.  This increase in receptive field size was done to compensate for higher GSD imagery, which assisted the model in learning local relationships between object classes. 

Second, the network capacity of each convolutional autoencoder (CAE) was decreased to compensate for the reduced dimensionality of RIT-18 (only six bands).  For each CAE, there are 32 units in the first convolution block, 64 units in the second convolution block, 128 units in the third, and 256 units in the $1 \times 1$ convolution.  The refinement blocks have the same number of units as their corresponding convolution block, so the last hidden layer of each CAE has 32 features. 

Third, as was done with MICA, the RBF-SVM classifier was replaced with a MLP classifier in order to speed up training.  An entire orthomosaic is passed through the SCAE network to generate three $N\times 32$ feature responses.  These feature responses are concatenated, convolved with a $5 \times 5$ mean-pooling filter, and then reduced to 99\% of the original variance using whitened principal component analysis (WPCA).  The final feature response is passed to a MLP classifier with the same architecture used by the MICA model in Section \ref{section:mica}.  

We also introduce a multi-resolution segmentation (MRS) method that uses SCAE features.  We combined object-based features with learned spatial-spectral features extracted from the SCAE deep learning framework. For object-based features, we hyper-segmented the orthomosaic images using the mean-shift clustering algorithm.  These orthomosaics were hyper-segmented at different scales to improve the segmentation of objects at different sizes.  Mean-shift was chosen because it automatically determines how many clusters should be in the output image.  The inputs to the mean-shift algorithm were the six spectral channels, the pixel location, and computed normalized-difference vegetation index (NDVI) for each pixel.  The object based features for the pixels in each cluster are the area and both spatial dimensions for each cluster.  We denote this method MRS+SCAE.

\section{RIT-18 Dataset}
\label{section:rit-18}

RIT-18 is a high-resolution (4.7 cm) benchmark, designed to evaluate the semantic segmentation of MSI, collected by a UAS.  UAS collection of non-RGB imagery has grown in popularity, especially in precision agriculture, because it is more cost effective than manned flights and provides better spatial resolution than satellite imagery.  This cost savings allows the user to collect data more frequently, which increases the temporal resolution of the data as well. The applications for UAS with MSI payloads include crop health sensing, variable-rate nutrient application prescription, irrigation engineering, and crop-field variability~\cite{huang2010multispectral}.

\subsection{Collection Site}

The imagery for this dataset was collected at Hamlin Beach State Park, located along the coast of Lake Ontario in Hamlin, NY.  The training and validation data was collected at one location, and the test data was collected at a different location in the park.  These two locations are unique, but they share many of the same class-types.  Table \ref{table:collection} lists several other collection parameters.

\begin{table}[t!]
\centering
\begin{tabular}{@{}lccc@{}}\toprule
 & \textbf{Train} & \textbf{Validation} & \textbf{Test} \\ \midrule 
 \textbf{Date}&29 Aug 16&6 Sep 16 &29 Aug 16\\
 \textbf{Time (UTC)}&13:37 &15:18 &14:49\\
 \textbf{Weather} & \multicolumn{3}{c}{Sunny, clear skies}\\
 \textbf{Solar Azimuth}& 109.65\degree
 & 138.38\degree &126.91\degree\\
 \textbf{Solar Elevation}& 32.12\degree & 45.37\degree &43.62\degree \\
\bottomrule \\
\end{tabular}
\caption{Collection parameters for training, validation, and testing folds for RIT-18.}
\label{table:collection}
\end{table}

\subsection{Collection Equipment}

\begin{figure}[t!]
\begin{center}
   \includegraphics[width=0.5\linewidth]{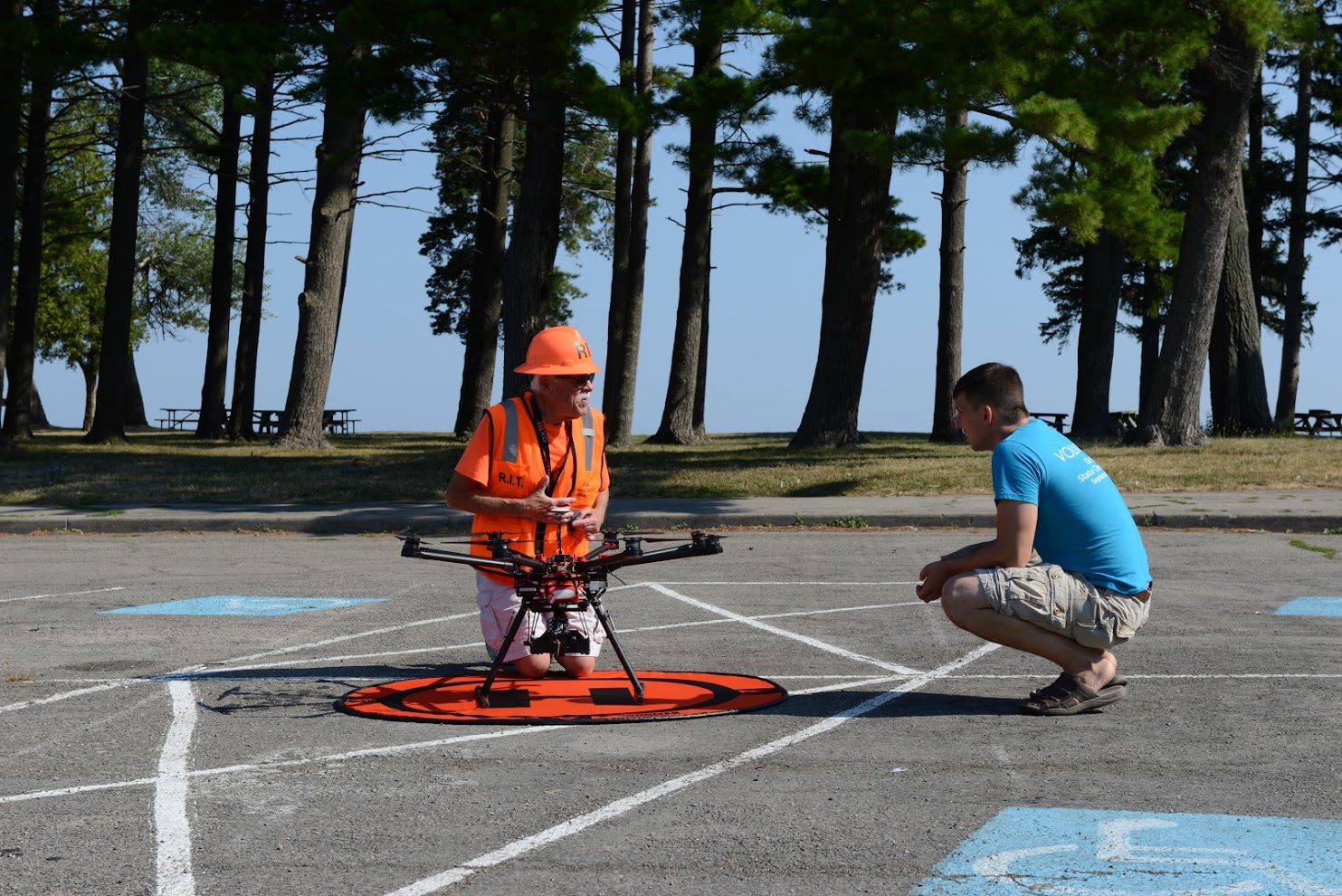}
\end{center}
   \caption{Tetracam Micro-MCA6 mounted on-board the DJI-S1000 octocopter prior to collection.}
\label{fig:octocopter}
\end{figure}

The equipment used to build this dataset and information about the flight is listed in Table \ref{table:specs}.  The Tetracam Micro-MCA6 MSI sensor has six independent optical systems with bandpass filters centered across the VNIR spectrum.  The Micro-MCA6 has been used on-board UASs to perform vegetation classification on orthomosaic imagery~\cite{laliberte2011multispectral} and assess crop stress by measuring the variability in chlorophyll fluorescence~\cite{ZarcoTejada20091262} and through the acquisition of other biophysical parameters~\cite{berni2009thermal}.  Fig. \ref{fig:octocopter} shows an image of the Micro-MCA6 mounted on-board the DJI-S1000 octocopter. 

\begin{table}[ht!]
\centering
\begin{tabular}{@{}ll@{}}\toprule
\multicolumn{2}{l}{\textbf{Imaging System}} \\
\midrule 
Manufacturer/Model & Tetracam Micro-MCA6\\
Spectral Range [nm] & 490-900\\
Spectral Bands & 6 \\
RGB Band Centers [nm] & 490/550/680 \\
NIR Band Centers [nm] & 720/800/900 \\
Spectral f [nm] & 10 (Bands 1-5)\\
                   & 20 (Band 6)\\
Sensor Form Factor [pix] & 1280x1024 \\
Pixel Pitch [$\mu$m] &  5.2 \\
Focal Length [mm]& 9.6\\
Bit Resolution & 10-bit \\
Shutter & Global Shutter \\
\midrule
\multicolumn{2}{l}{\textbf{Flight}} \\ \midrule
Elevation [m] & 120 (AGL) \\
Speed [m/s] & 5 \\
Ground Field of View [m] & $\approx$60x48\\
GSD [cm] & 4.7\\
Collection Rate [images/sec] & 1 \\
\bottomrule \\
\end{tabular}
\caption{Data Collection Specifications}
\label{table:specs}
\end{table}

\subsection{Dataset Statistics and Organization}

The RIT-18 dataset (Fig. \ref{fig:ortho}) is split up into training, validation, and testing folds. Each fold contains an orthomosaic image and corresponding classification map.  Each orthomosaic contains the six-band image described in Section \ref{section:ortho} along with a mask where the image data is valid.  The spatial dimensionality of each orthomosaic image  is 9,393$\times$5,642 (train), 8,833$\times$6,918 (validation), and 12,446$\times$7,654 (test).

\begin{figure}[t!]
\begin{center}
   \includegraphics[width=0.9\linewidth]{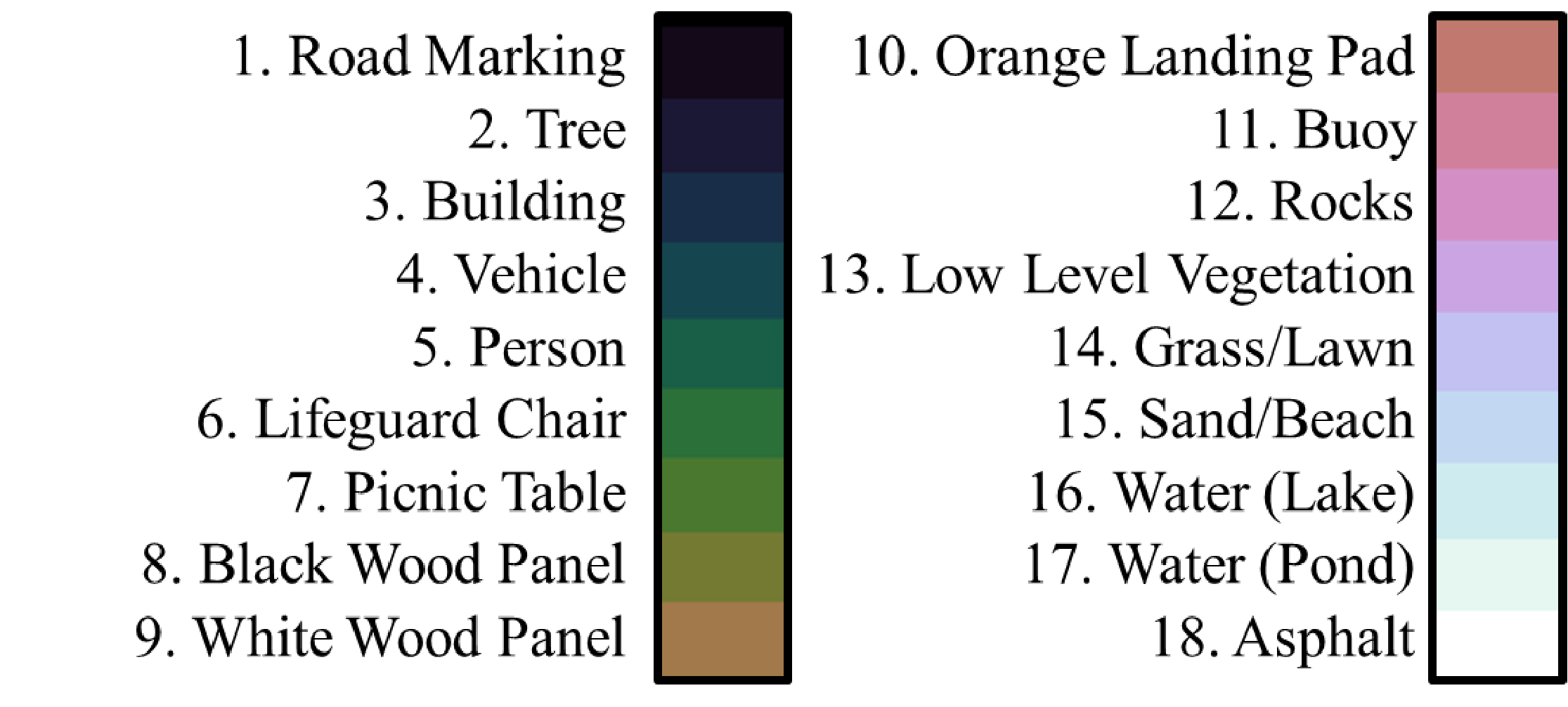}
\end{center}
   \caption{Class labels for RIT-18.}
   \label{fig:colormap}
\end{figure}

Figure~\ref{fig:colormap} lists the 18 class labels in RIT-18 and their corresponding color map (used throughout this paper).  Each orthomosaic was hand-annotated using the region-of-interest (ROI) tool in ENVI.  Several individuals took part in the labeling process.  More information about these classes can be found in \ref{section:classes}.

\begin{figure}[t!]
\begin{center}
   \includegraphics[width=0.9\linewidth]{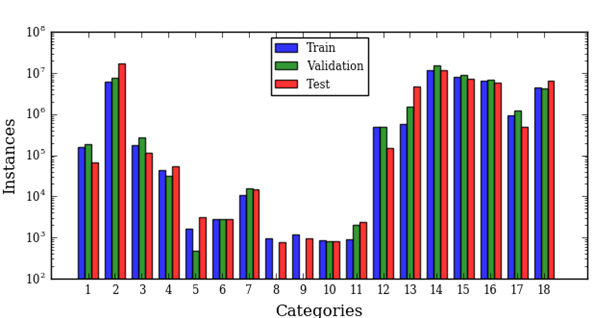}
\end{center}
   \caption{Class-label instances for the RIT-18 dataset.  Note: The y-axis is logarithmic to account for the number disparity among labels.}
\label{fig:stats}
\end{figure}

The class-labeled instances are, as illustrated in Fig. \ref{fig:stats}, orders of magnitude different from one-another.  These underrepresented classes should make this dataset more challenging.  

\section{FCN Training Procedures}
\label{section:net-training}

\subsection{Pre-Training the ResNet-50 DCNN on Synthetic MSI}
\label{section:initialization}
The two FCN models are assessed with and without DCNNs that are pre-trained on synthetic DIRSIG imagery. For the pre-trained model, the ResNet-50 DCNN was initialized using 80$\times$80 pixel patches, which is approximately 4.7 million DIRSIG training images.  The network was trained using a mini-batch size of 128 and a weight-decay of 1e-4.  The weights were randomly initialized from a zero-mean normal distribution.  The labels provided by DIRSIG are a class-label for each pixel; however, the ResNet-50 DCNN is trained to perform image classification.  We assigned a label to each image patch based on the most common label.

We compute the channel-mean and standard-deviation using the entire DIRSIG training set, and these parameters were used to scale each image to zero-mean/unit-variance.  During training, the images are shuffled for each epoch, and we use random horizontal and vertical flips for data augmentation.  Because the class distribution for the DIRSIG data is unbalanced, we use the class-weights $w_i$ for the entire training set to build sample (per-pixel) weights.   The class-weights were assigned using Equation \ref{eq:sample_weights} with $\mu=0.15$.  We optimized the network using Nadam with an initial learning rate of 2e-3.  We then dropped the learning rate when the validation loss plateaued. 

\subsection{DCNN Fine-Tuning}
\label{section:finetune}

We trained both semantic segmentation frameworks using randomly initialized DCNNs and with the DCNN pre-trained on the DIRSIG dataset.  Because of the high-resolution of each orthomosaic, the data is broken into $160 \times 160$ patches and fed to the segmentation frameworks. The randomly initialized models are trained end-to-end in one stage because the filters of its DCNN need to be tuned. All weights that do not come from the pre-trained DCNN are randomly initialized from a zero-mean Gaussian distribution. For the models that did not use pre-training, an initial learning rate of  2e-3 is used, and then the learning rate is dropped by a factor of 10 four times as the validation loss plateaued.

For the models that use the pre-trained DCNN, we train them in two stages. First, the pre-trained portion of the model is frozen and the remaining layers are trained to adapt the weights of the segmentation network to the pre-trained weights in the convolution network. An initial learning rate of 2e-3 is used for this stage, and then it is dropped by a factor of 10 when validation loss plateaus.  Second, we fine-tune both the pre-trained portion and the segmentation network jointly using an initial learning rate of 2e-5; and again, drop it by a factor of 10 four times as the validation loss plateaued. 

\sloppy
All models are optimized using the Nadam optimizer, batch size of 32, weight decay of 1e-4, and class-weight parameter of $\mu=0.25$, which optimizes for AA.  Plots for the training and validation accuracy and loss, for both FCN models, are provided in \ref{appendix:results}.   We observed that the pre-trained models, especially RefineNet, saturated at the peak accuracy more quickly (fewer epochs) than the randomly initialized networks because most of the weights are close to their final solution.  The fine-tuned models continue to make minor improvements over a longer period of time because it employs a lower learning rate to prevent from completely overwriting the pre-trained DIRSIG weights; whereas, the random weight initialization stops training more quickly because it is not attempting to preserve any pre-trained weights.

\section{Experimental Results}
\label{section:benchmarks}

\subsection{RIT-18 Results}

Results for all algorithms on the RIT-18 test set are listed in Table \ref{table:results}. The table shows the classification performance of our DCNN segmentation models, with (Sim) and without (Rdm) initializing the network using synthetic data. Pretraining the ResNet-50 DCNN on the synthetic imagery required  three weeks on our GPU.  The number of batch updates performed which was higher than the number performed in [35].  The network fine-tuning operation took 17.1 hours for Sharpmask and 56.7 hours for RefineNet.  Each algorithm is evaluated on per-class accuracy and mean-class accuracy (AA).  AA is used as our primary metric due to the disparity in class distribution, and all algorithms except kNN were trained to compensate for class unbalance.  Random chance for assigning a class to a given pixel is 5.6\%. Models that perform poorly on some classes, especially classes that perform worse than chance, have over-fit to perform better on classes with more training samples.  

When synthetic image initialization was used, both Sharpmask and RefineNet outperform all of the other algorithms in mean-class accuracy performance, demonstrating that advanced techniques used in computer vision can be effectively used with remote sensing MSI data. For mean-class accuracy, the best performing model is RefineNet-Sim.  Both of the FCN models that do not use pre-training perform better on classes with more samples, but the mean-class accuracy shows that these models are not as discriminative as their counterparts that are initialized with synthetic data. RefineNet-Rdm overfit to the classes with the most samples, whereas RefineNet-Sim was more discriminative with the pre-trained ResNet-50 weights. 

Sharpmask-Rdm performed only slightly worse than Sharpmask-Sim, whereas RefineNet-Sim's results were far greater than RefineNet-Rdm. This discrepancy is likely because RefineNet has 5.8 times as many trainable parameters compared to Sharpmask. Sharpmask is a relatively shallow semantic segmentation framework, it only uses the early layers of ResNet-50, and it only has 11.9 million trainable parameters. In comparison, RefineNet has 69 million trainable parameters. The more parameters a model has the greater its capacity, but more parameters also mean more labeled data is needed to prevent overfitting. We suspect that if a deeper pre-trained network was used, e.g., ResNet-152, then we would see even greater differences between pre-trained and randomly initialized models. 

Comparing the baseline models, we see that MICA yielded a 4.8 percent increase in mean-class accuracy from the simpler MP experiment, demonstrating that unsupervised feature extraction can boost classification performance for high-resolution imagery. Unlike earlier work~\cite{Kemker16}, MICA outperformed SCAE showing that low-level features over a larger receptive field could be more important than higher-level features over a smaller spatial extent.  MRS+SCAE improved performance over SCAE for objects where the confusion between class predictions was based on the object's size, which means that object based methods, such as CRF or GEOBIA, could increase segmentation performance in future frameworks, especially when the dominant source of error is salt-and-pepper label errors.

The main issue with unsupervised feature extraction methods like SCAE is that they are unable to learn object-specific features without image annotations and will instead learn the dominant spatial-spectral features present in the training imagery (i.e., they may miss small objects and other classes not commonly present).  In contrast, supervised DCNN frameworks, if properly trained and regularized, can better learn the spatial-spectral feature representation for each object class; however, these models need to be pre-trained with large annotated datasets to generalize well in the first place.  This is why the end-to-end FCN frameworks pre-trained with synthetic imagery is important for the remote sensing community.  It is entirely possible to train these FCN models with object based features in the end-to-end framework, but this comes with an additional computation burden in contrast to end-to-end FCN models that \emph{learn} to sharpen object boundaries.

\begin{table}[t!]
\scriptsize
\centering
\setlength{\tabcolsep}{.25em}
\begin{tabular}{@{}r|rrrrrrrrrrr@{}}\toprule
 \multirow{ 2}{*}{} & \multirow{ 2}{*}{\textbf{kNN}} & \multirow{ 2}{*}{\textbf{SVM}} & \multirow{ 2}{*}{\textbf{MLP}} & \multirow{ 2}{*}{\textbf{MP}} &  \multirow{ 2}{*}{\textbf{MICA}} & \multirow{ 2}{*}{\textbf{SCAE}} & \textbf{MRS+} & \multicolumn{2}{c}{\textbf{Sharpmask}} & \multicolumn{2}{c}{\textbf{RefineNet}} \\ 
& & & & & & & \textbf{SCAE} &\multicolumn{1}{c}{\textbf{Rdm}} & \multicolumn{1}{|c}{\textbf{Sim}}  & \textbf{Rdm} & \multicolumn{1}{|c}{\textbf{Sim}} \\ \midrule 
\textbf{Road Markings}   & 65.1 & 51.0 & 75.6 & 29.6 & 43.2 & 37.0&63.3 &78.3 &\textbf{92.5}&0.0 &59.3 \\
\textbf{Tree}            & 71.0 & 43.5 & 62.1 &44.1 & 92.6 & 62.0& 90.5&\textbf{93.0} &89.8&91.0 &89.8 \\
\textbf{Building}        & 0.3 & 1.5 & 3.7 &0.6 & 1.0 & 11.1&0.7 &6.3  &16.5 &0.0 &\textbf{17.0} \\
\textbf{Vehicle}         & 15.8 & 0.2 & 1.0 &0.2 & 47.5 & 11.8&53.6&51.7 &20.8&0.0 &\textbf{61.9} \\
\textbf{Person}          & 0.1 & 19.9 & 0.0 & 31.2 & 0.0 &  0.0&2.3 &29.9 & 14.9 &0.0 &\textbf{36.8} \\
\textbf{Lifeguard Chair} & 1.0 & 22.9 & 3.1 & 16.9 & 50.8 & 29.4&2.8 &44.5 &\textbf{85.1}&0.0 &80.6 \\
\textbf{Picnic Table}    & 0.6 & 0.8 & 0.0 & 0.6 & 0.0 & 0.0&30.3 &11.4 &32.9&0.0 &\textbf{62.4} \\
\textbf{Black Panel}     & 0.0 & \textbf{48.3} & 0.0 & 47.9 & 0.0 & 0.0&18.9 &0.0  &0.0 &0.0 &0.0 \\
\textbf{White Panel}     & 0.1 & 0.3 & 0.0 & 0.8 & 0.0 & 0.4&0.0 &4.6 &9.8&0.0 &\textbf{47.7} \\
\textbf{Orange Pad}      & 14.6 & 15.2 & 77.4 & 22.1 & 66.3 & 99.3 &0.0 & \textbf{100.0} & 97.1&0.0 &\textbf{100.0 }\\
\textbf{Buoy}            & 3.6 & 0.7 & 1.8 & 10.1 & 0.0 & 7.2&55.6 &71.0 &82.4&0.0 &\textbf{85.8} \\
\textbf{Rocks}           & 34.0 & 20.8 & 38.8 & 33.4 & 66.5 & 36.0&42.8 &79.0 &\textbf{87.8}&87.3&81.2 \\
\textbf{Low Vegetation}  & 2.3 & 0.4 & 0.3 & 0.1 & 13.3 & 1.1&5.1 &\textbf{ 22.9} &0.5 &0.0 &1.2 \\
\textbf{Grass/Lawn}      & 79.2 & 71.0 & 85.4 & 73.1 & 84.8 & 84.7&83.6 &84.8 &87.4&88.4&\textbf{90.3} \\
\textbf{Sand/Beach}      & 56.1 & 89.5 & 36.4 & \textbf{95.2} & 78.2 & 85.3 &8.0 &73.4&91.2&6.8 &92.2 \\
\textbf{Water (Lake)}    & 83.6 & 94.3 & 92.6 & 94.6 & 89.1 &  \textbf{97.5}& 90.0 &96.2 &93.3&90.4&93.2 \\
\textbf{Water (Pond)}    & 0.0 & 0.0 & 0.0 & 0.2 & \textbf{3.4} & 0.0&0.0 &0.0  &0.0 &0.0 &0.0 \\
\textbf{Asphalt}         & 80.0 & 82.7 & 93.1 & 93.3 & 46.9 & 59.8& 72.0&\textbf{96.2} &92.1&95.9&77.8 \\
\midrule 
\textbf{AA}              & 27.7 & 29.6 & 30.4 & 31.3 & 36.2 & 32.1& 34.4&52.4 &57.3& 30.1&\textbf{59.8} \\
\bottomrule 

\end{tabular}
\caption{Per-class accuracies as well as mean-class accuracy (AA) on the RIT-18 test set. The two initializations used for our Sharpmask and RefineNet models include random initialization (Rdm) and a network initialized with synthetic data (Sim).  We compare our results against the benchmark classification frameworks listed in Section \ref{section:benchmarks}.}
\label{table:results}
\end{table}

\begin{figure}[t]
    \centering
    \subfigure{%
     	\includegraphics[width=0.23\linewidth]{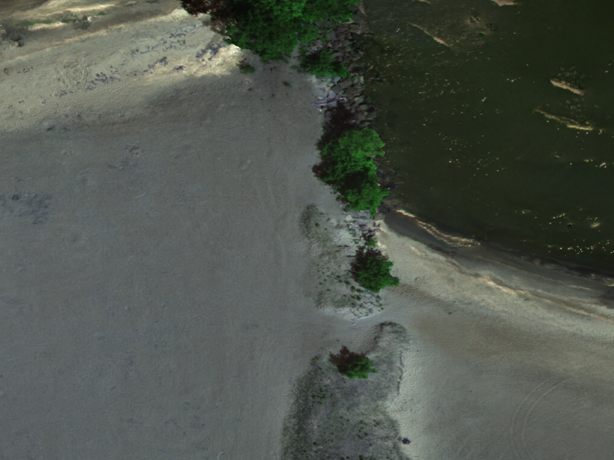} 	 }
     \subfigure{%
     	\includegraphics[width=0.23\linewidth]{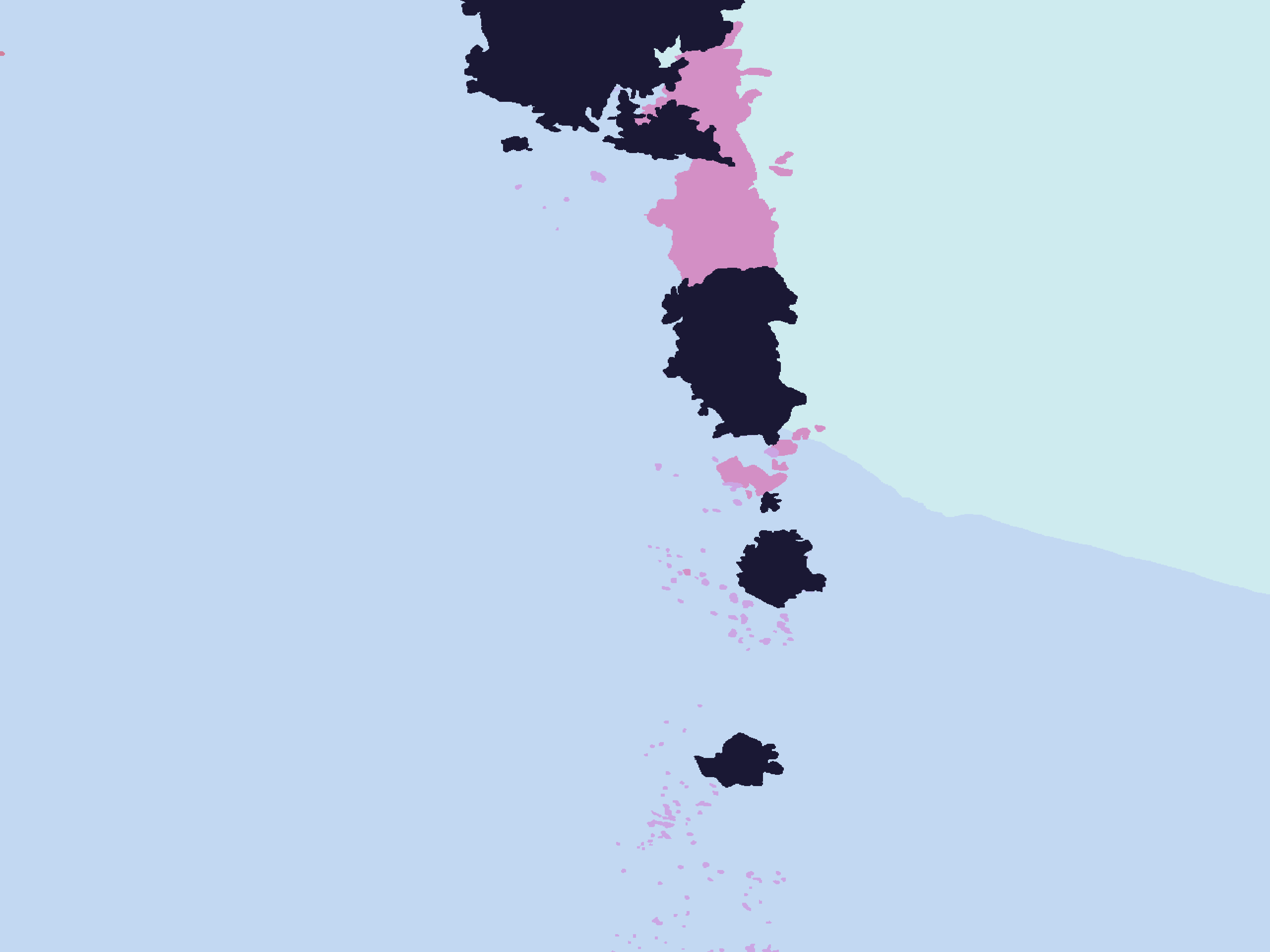} 	 }
     \subfigure{%
     	\includegraphics[width=0.23\linewidth]{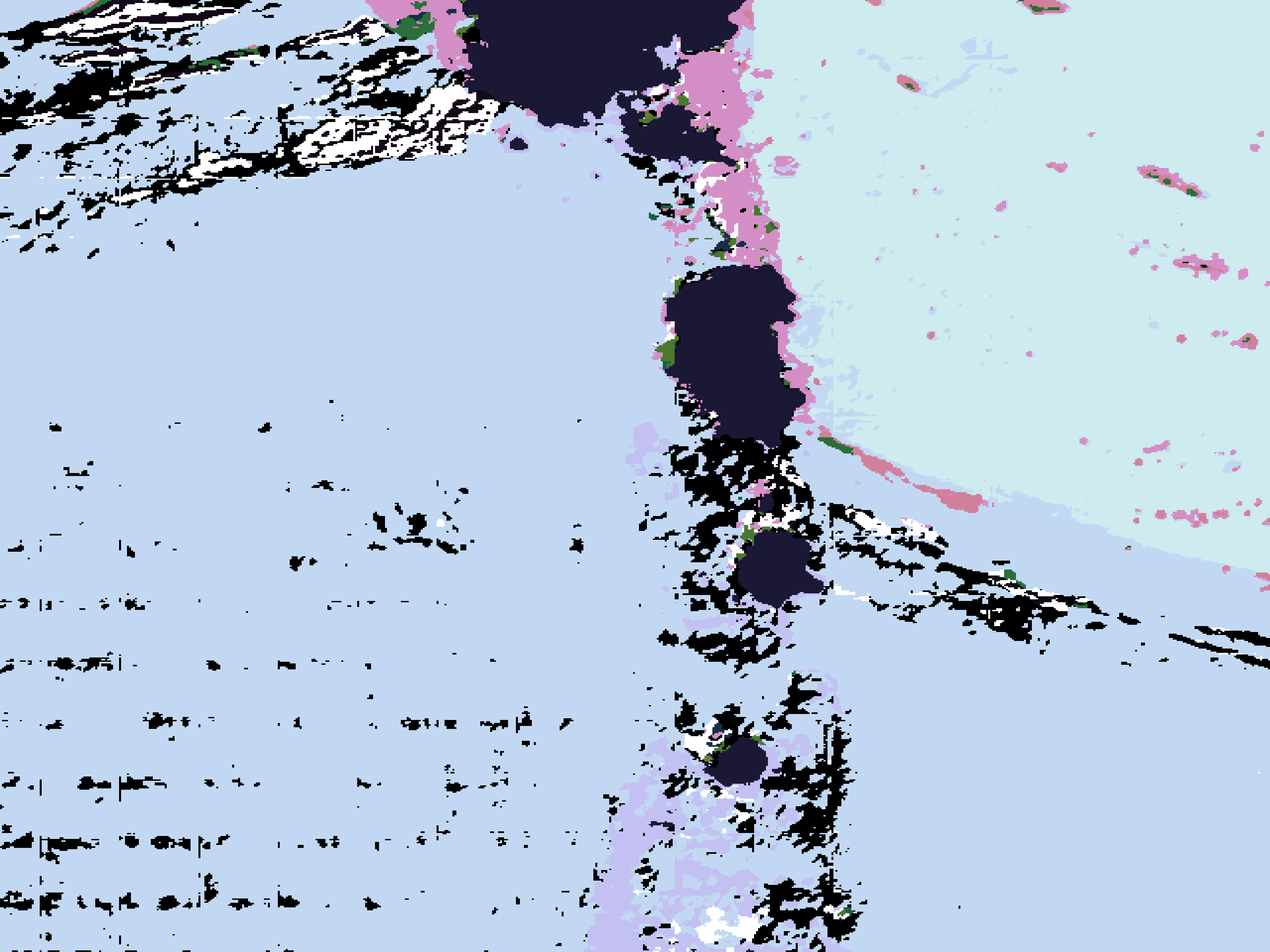} 	 }
     \subfigure{%
     	\includegraphics[width=0.23\linewidth]{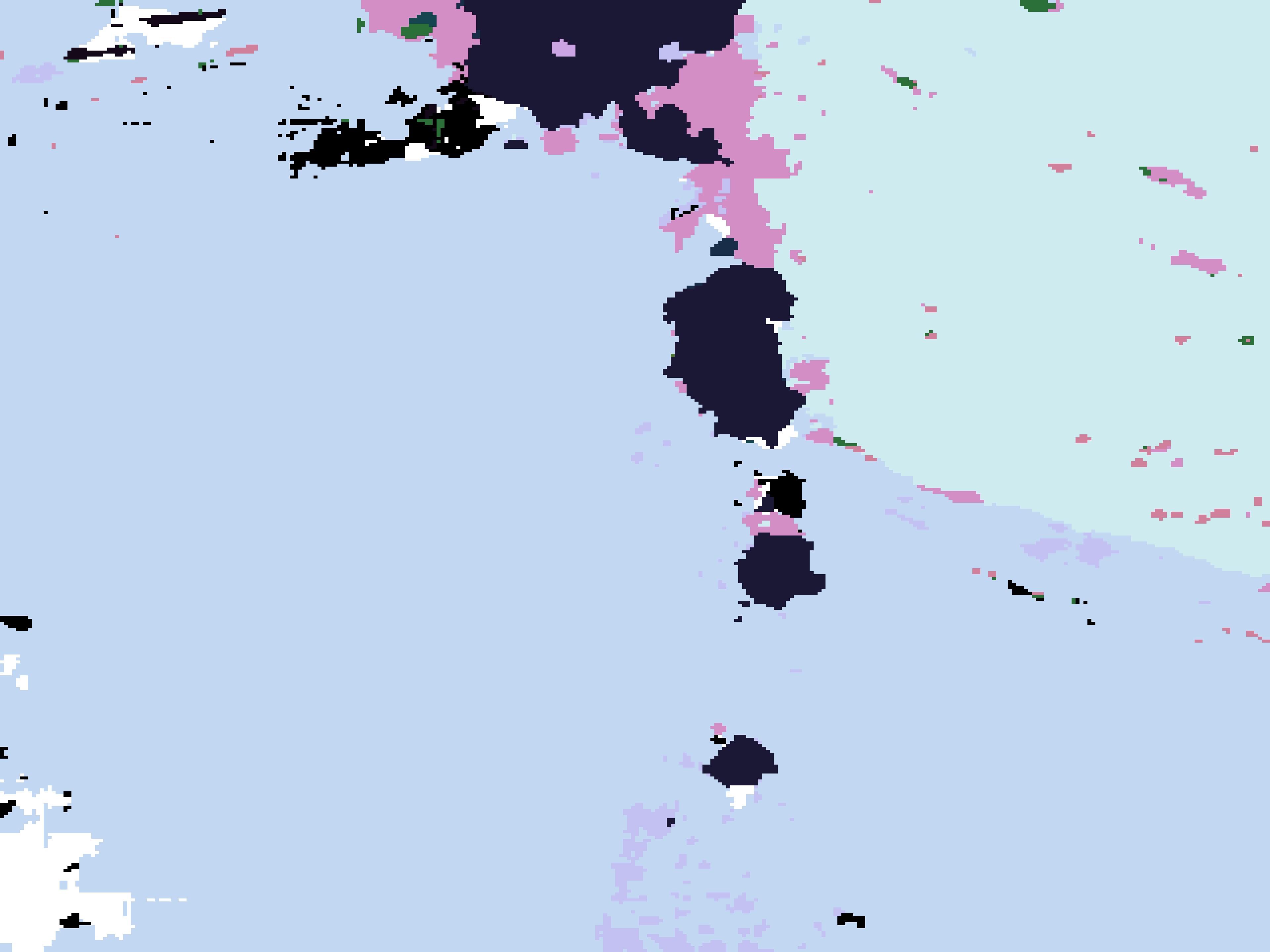} 	 }        
    \subfigure{%
     	\includegraphics[width=0.23\linewidth]{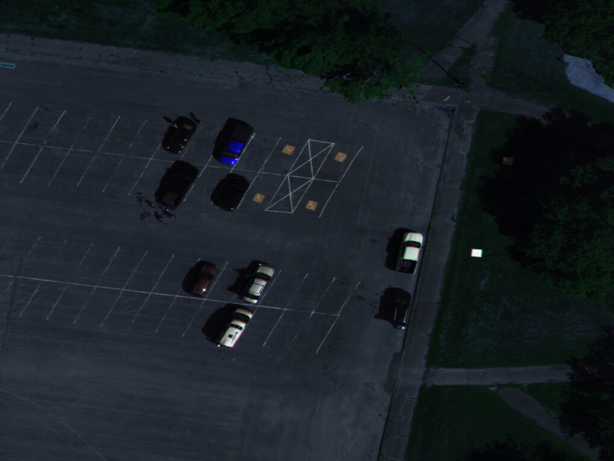} 	 }
     \subfigure{%
     	\includegraphics[width=0.23\linewidth]{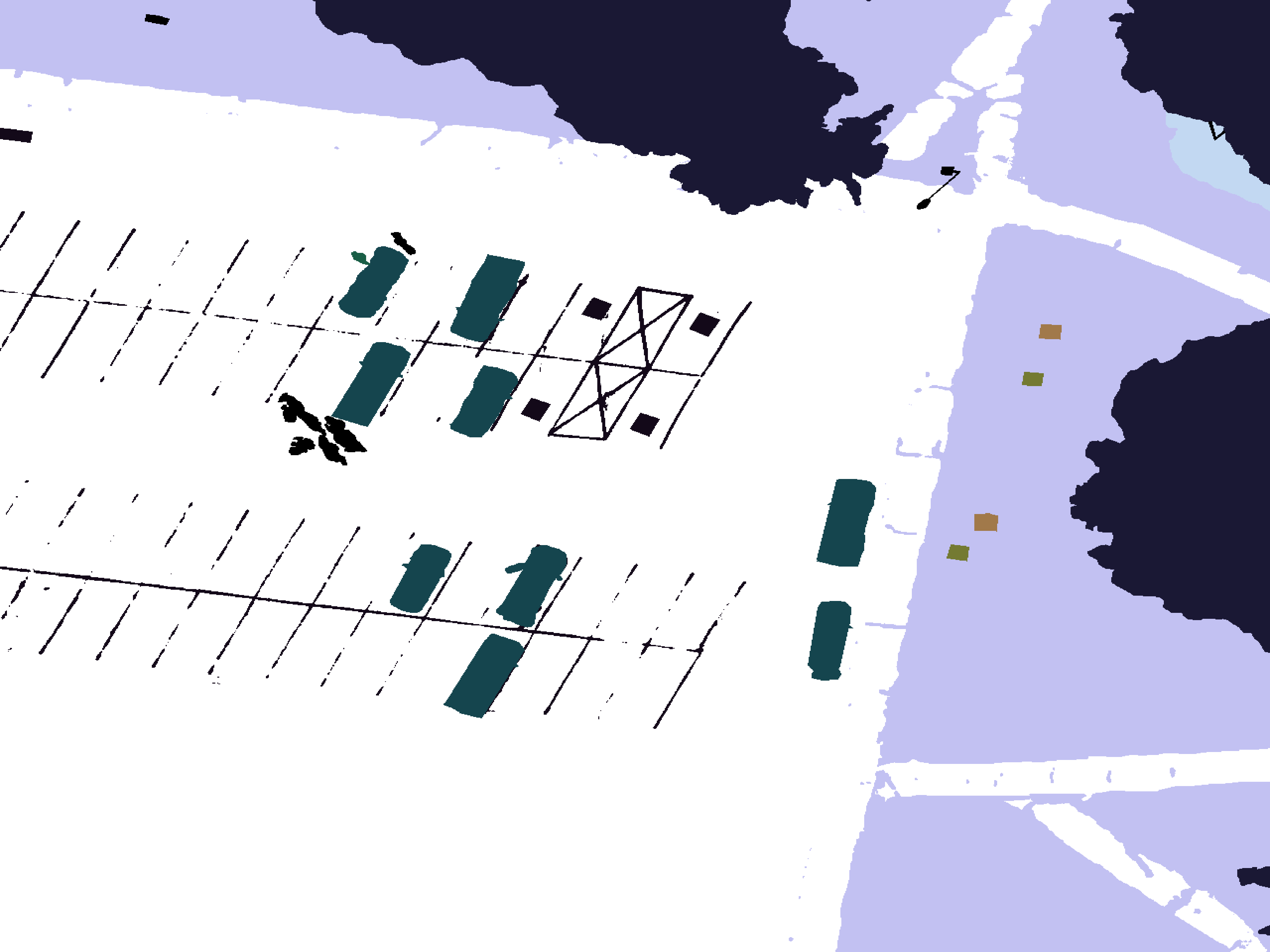} 	 }
     \subfigure{%
     	\includegraphics[width=0.23\linewidth]{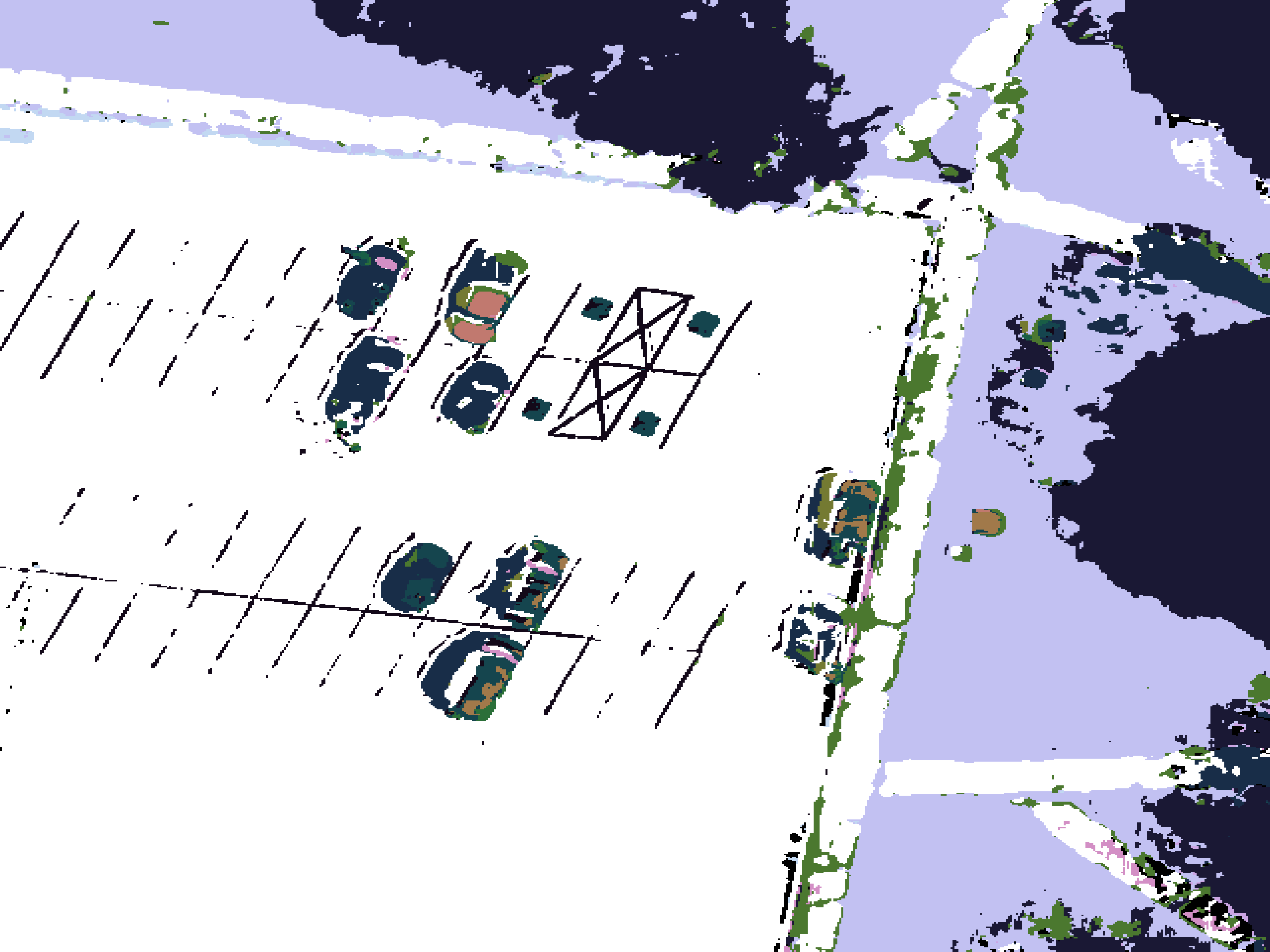} 	 }
     \subfigure{%
     	\includegraphics[width=0.23\linewidth]{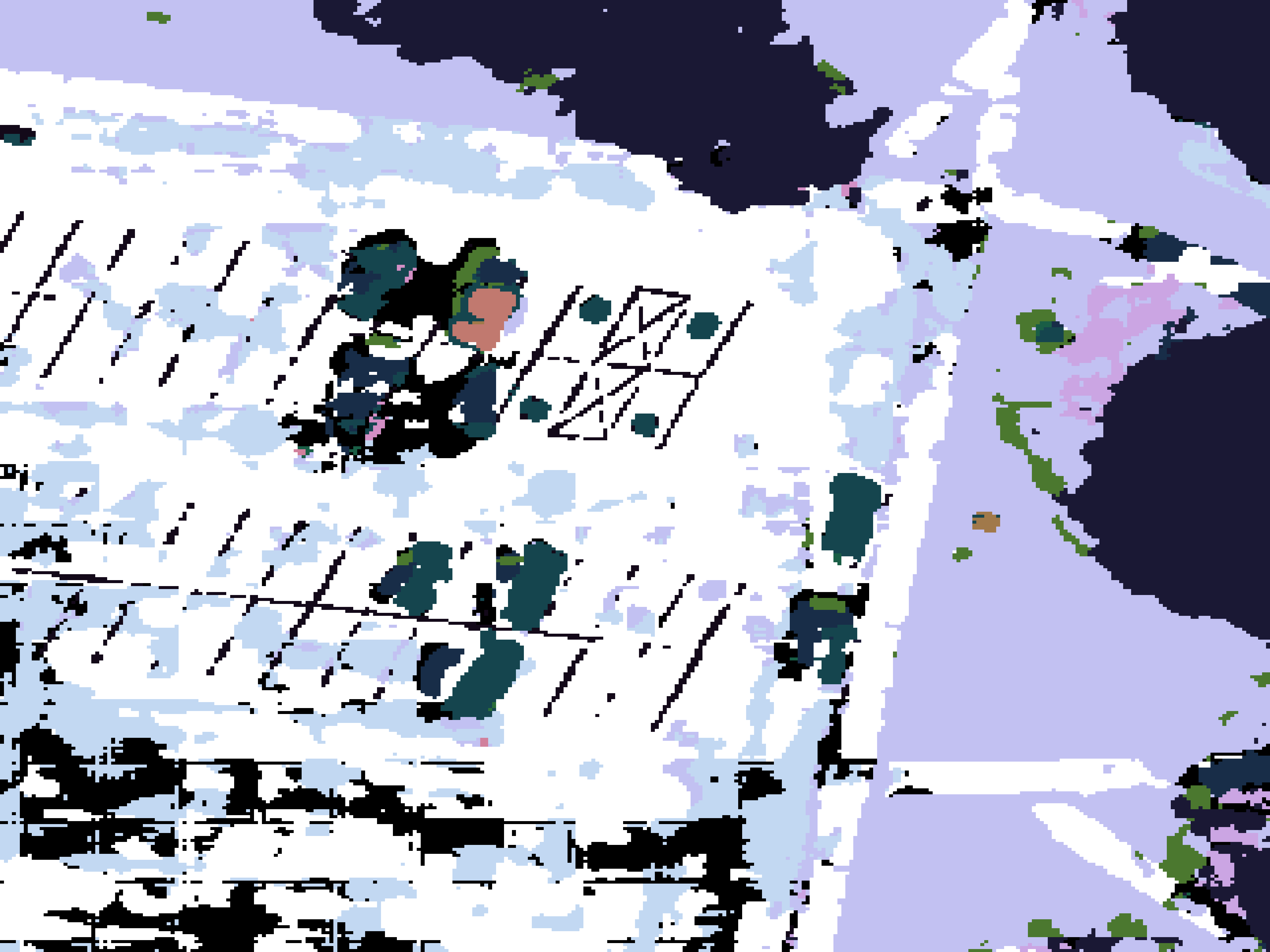} 	 }   
     \addtocounter{subfigure}{-8}
     \subfigure[Test Image]{%
     	\includegraphics[width=0.23\linewidth]{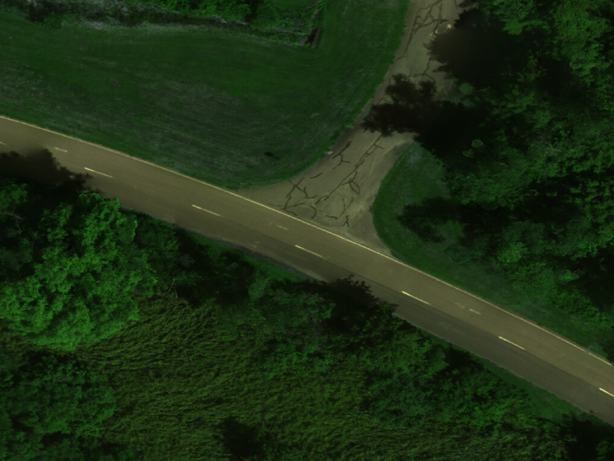} }
     \subfigure[Ground Truth]{%
     	\includegraphics[width=0.23\linewidth]{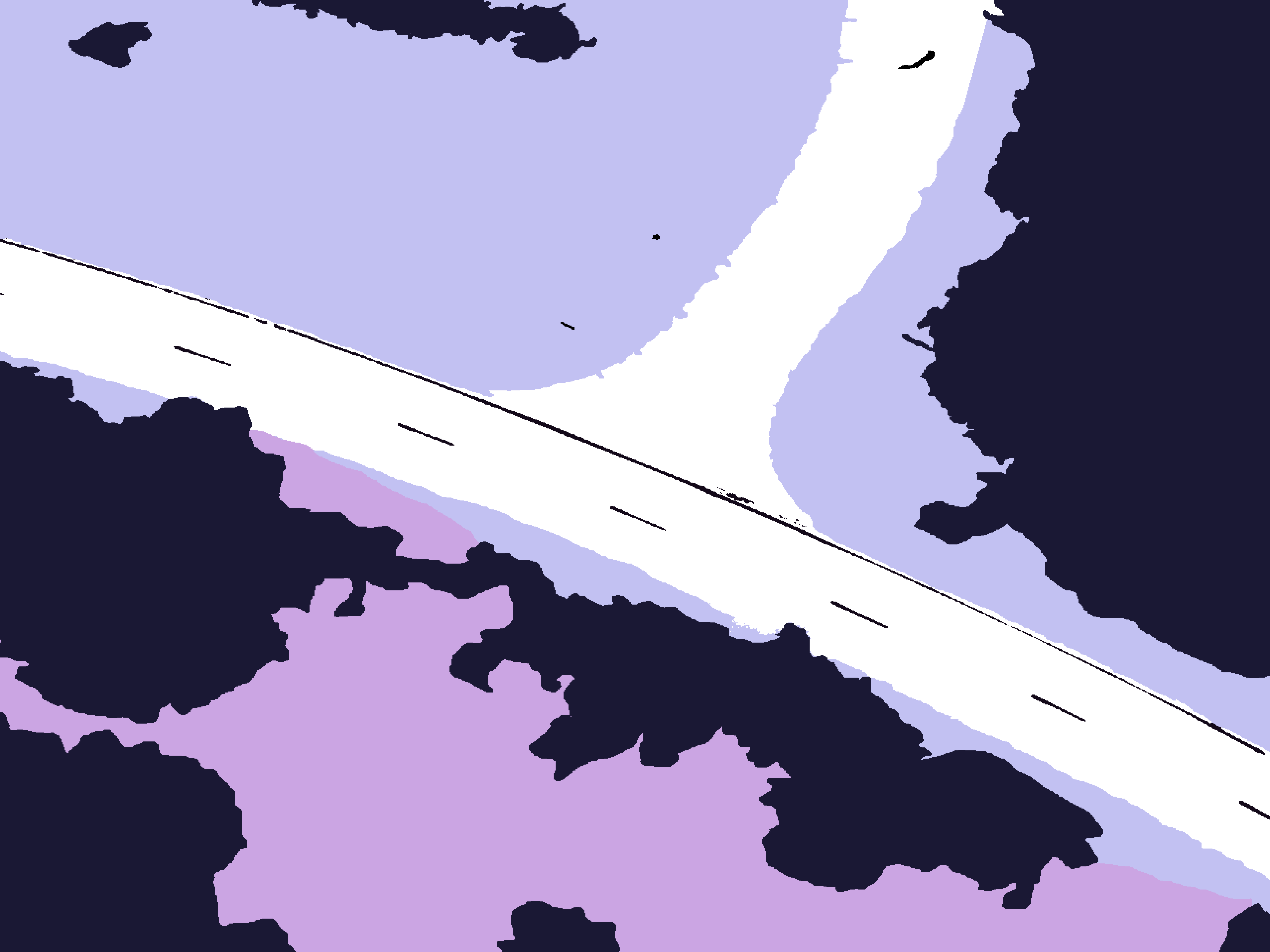} }
     \subfigure[Sharpmask]{%
     	\includegraphics[width=0.23\linewidth]{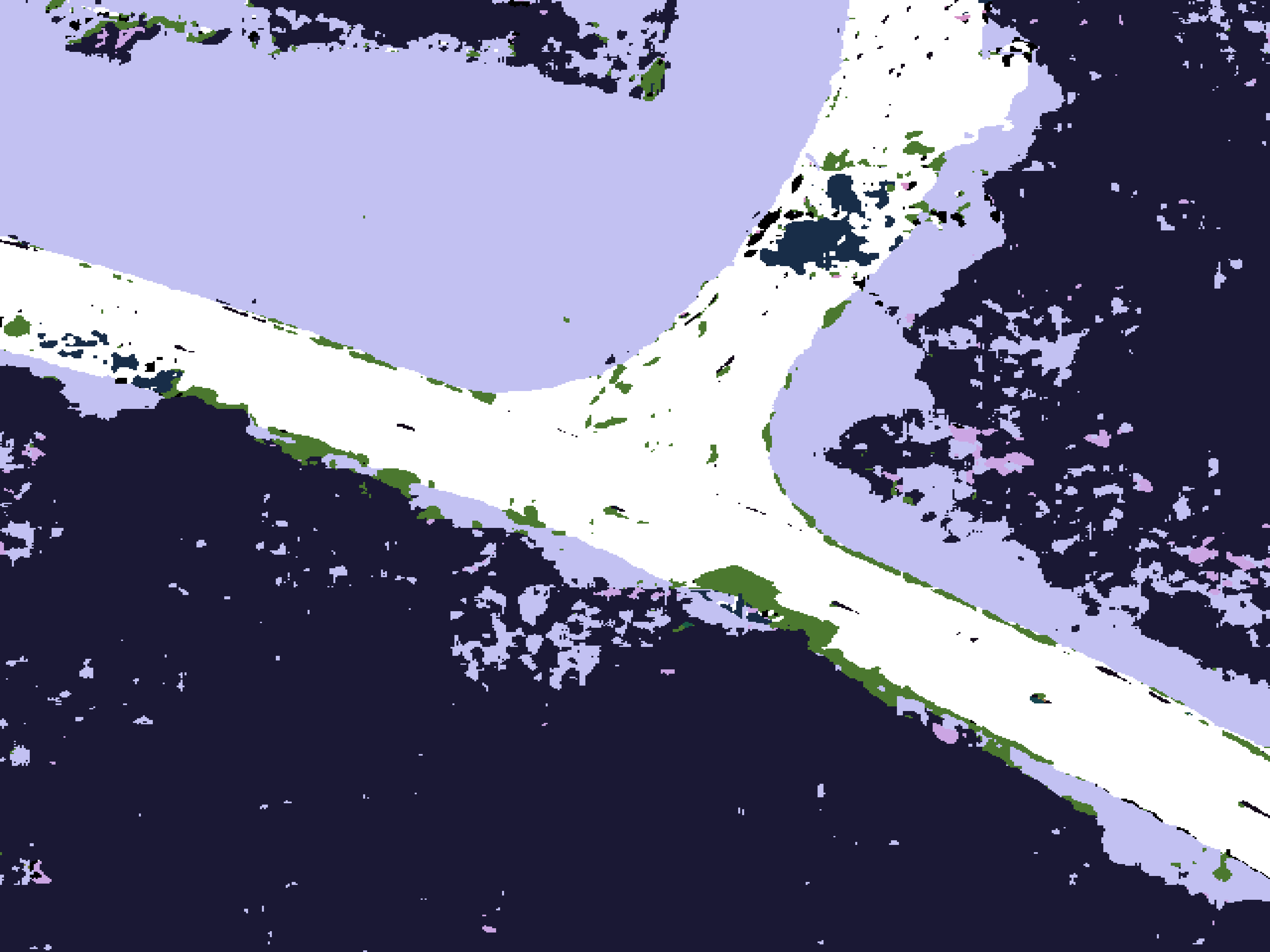} }
   \subfigure[RefineNet]{%
     	\includegraphics[width=0.23\linewidth]{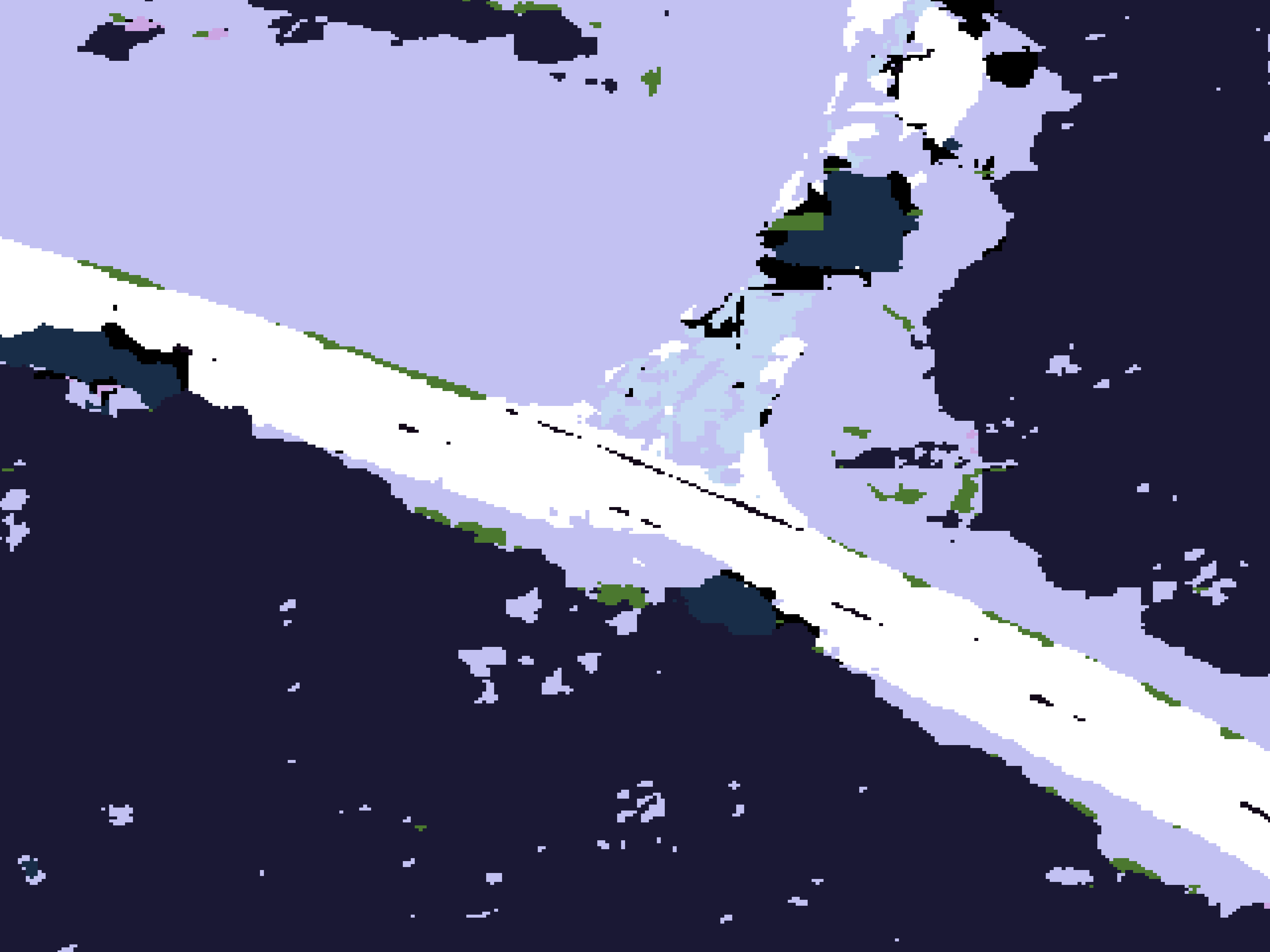} }  
    \caption{Experimental results for Sharpmask and RefineNet models.  These images are small patches taken from the test orthomosaic.}
    \label{fig:results}
\end{figure}

The FCN models failed to classify the black panel, low-level vegetation, and the pond.  According to the confusion matrices (\ref{appendix:results}), the black-panel was predominantly misclassified as asphalt, likely because 1) black-panel had very few training samples and 2) they shared similar spatial/spectral characteristics in the VNIR spectrum.  The low-level vegetation was misclassified as trees and the pond water was misclassified as grass/lawn for similar reasons.  Large portions of the pond, especially in the test image, contain some vegetation in and on-top of the water which is why it could have been confused for grass.  A possible solution to correcting this problem is to initialize the DCNN with DIRSIG imagery that replicates these conditions (e.g.,  vegetation in a body of water).

Fig.~\ref{fig:results} shows a sample of the predictions made by our Sharpmask-Sim and RefineNet-Sim frameworks.  Sharpmask does a better job at classifying the road, road markings, and vehicles; and RefineNet had fewer classification artifacts over the beach area.  This shows that certain model architectures can be robust to illumination invariance in rough surfaces caused by BRDF effects.  RefineNet is likely more robust because it extracts features from deeper convolutional layers and the chained residual pooling aids in capturing background context.  Both models did a good job classifying the grass and, for the most part, the lake; however, the low-level vegetation area seemed to be mis-classified as trees - which has orders of magnitude more training samples.  Sharpmask and RefineNet would also tend to mis-classify the parts of Lake Ontario, where the wave-crest is whiter than the rest of the body of water, as rocks.  This is likely because the rocks in RIT-18 are dominantly surrounded by darker lake water, so the DCNN models have been trained to associate the relatively brighter patches in the lake as rocks.  This could be remedied during training by revisiting areas in the lake that have a higher spatial variability or collecting additional examples where the white wave-crests occur.  

\subsection{Band Analysis for RIT-18}
\label{section:band_analysis}

The focus of this paper is on semantic segmentation of MSI data using FCN models with and without pre-training, but how useful are the non-RGB bands? To assess this, we conducted two experiments using RIT-18. First, we trained the SVM model on different channels. This analysis, shown in Table \ref{table:band_selection}, includes only the RGB bands (SVM-RGB), only the three NIR bands (SVM-NIR), a false-color image (SVM-CIR), and a four band RGB-NIR (SVM-VNIR).  The 720 nm band was used for the SVM-CIR and SVM-VNIR experiments.  

\begin{table}[t!]
\centering
\setlength{\tabcolsep}{.25em}

\begin{tabular}{lr}\toprule
&   \textbf{AA} \\ \midrule
\textbf{SVM-RGB}  & 19.9 \\
\textbf{SVM-NIR}  & 26.0  \\
\textbf{SVM-CIR}  & 25.3 \\
\textbf{SVM-VNIR}  & 25.7  \\
\textbf{SVM} &   \textbf{29.6}  \\
\bottomrule \\
\end{tabular}

\caption{The effect of band-selection on  mean-class accuracy (AA).  For comparison, the last entry is the experiment from Table \ref{table:results} which used all six-bands.}
\label{table:band_selection}
\end{table}

This analysis suggests that the additional NIR bands have a large impact on performance; consistent with the fact that most of the scene is vegetation.  To further test this hypothesis, we pre-trained ResNet-50 using 4-band DIRSIG data, and then it was used to fine-tune RefineNet on the first four spectral channels of RIT-18. The results of the 4-band model compared to the full 6-band model are shown in Table~\ref{table:band_selection2}. These results indicate  that the Micro-MCA6 increase classification performance compared to simpler 4-band MSI systems.  The 4-band solution overfits to the dominant classes in RIT-18. This type of sensor could, in the future, provide an option for the fine-grained classification of various plant life.

\begin{table}[t!]
\footnotesize
\centering
\setlength{\tabcolsep}{.25em}
\begin{tabular}{@{}r|rr@{}}\toprule
 & \textbf{4-Band} & \textbf{6-Band} \\ \midrule 
\textbf{Road Markings}   &0.0  &\textbf{59.3} \\
\textbf{Tree}            &\textbf{95.8}  &89.8 \\
\textbf{Building}        &0.0  &\textbf{17.0} \\
\textbf{Vehicle}         &0.0  &\textbf{61.9} \\
\textbf{Person}          &0.0  &\textbf{36.8} \\
\textbf{Lifeguard Chair} &0.0  &\textbf{80.6} \\
\textbf{Picnic Table}    &0.0  &\textbf{62.4} \\
\textbf{Black Panel}     &0.0  &0.0 \\
\textbf{White Panel}     &0.0  &\textbf{47.7} \\
\textbf{Orange Pad}      &0.0  &\textbf{100.0 }\\
\textbf{Buoy}            &0.0  &\textbf{85.8} \\
\textbf{Rocks}           &0.0  &\textbf{81.2} \\
\textbf{Low Vegetation}  &0.0  &\textbf{1.2} \\
\textbf{Grass/Lawn}      &82.8  &\textbf{90.3} \\
\textbf{Sand/Beach}      &18.4  &\textbf{92.2} \\
\textbf{Water (Lake)}    &84.0  &\textbf{93.2} \\
\textbf{Water (Pond)}    &0.0  &0.0 \\
\textbf{Asphalt}         &6.9  &\textbf{77.8} \\
\midrule 
\textbf{AA}              &15.2  &\textbf{59.8} \\
\bottomrule 

\end{tabular}
\caption{Performance of RefineNet-Sim on the RIT-18 test set using 4-band (VNIR) and the full 6-band images.  These results include per-class accuracies and mean-class accuracy (AA).
}
\label{table:band_selection2}
\end{table}

\section{Discussion}

In this paper, we demonstrated the utility of FCN architectures for the semantic segmentation of remote sensing MSI.  An end-to-end segmentation model, which uses a combination of convolution and pooling operations, is capable of learning global relationships between object classes more efficiently than traditional classification methods.  Table \ref{table:results} showed that an end-to-end semantic segmentation framework provided superior classification performance on fourteen of the eighteen classes in RIT-18, demonstrating that the learned features in supervised DCNN frameworks are more discriminative than features built from unsupervised learning methods.

We showed that generated synthetic imagery can be used to effectively initialize DCNN architectures to offset the absence of large quantities of annotated image data.  Models that were initialized randomly showed degraded mean-class accuracy, a good metric for datasets with unbalanced class distributions.

DIRSIG could be used to generate large custom datasets for other imaging modalities such as multispectral, hyperspectral, LIDAR, or a combination of all the above.  These types of scenes could be developed thanks to the increase in UAS collection of remote sensing data. Our work could be adapted to these sensors. Evolving from multispectral to hyperspectral data will likely require modifications to our current models in order to deal with the higher dimensionality.  Initializing networks using DIRSIG could improve, not only semantic segmentation, but also networks for other tasks, such as object detection or target tracking in hyperspectral imagery.

Finally, we introduced the RIT-18 dataset as a benchmark for the semantic segmentation of MSI.  This dataset benefits from higher spatial resolution, large number of object classes, and wider spectral coverage to improve performance in vegetation-heavy scenes.  Datasets that are built from UAS platforms could be more practical for any commercial or research purpose.  The orthomosaic generation pipeline provided in \ref{section:ortho} could be utilized to quickly generate remote sensing products with little intervention. 

The absolute accuracy of the orthomosaic images in RIT-18 are limited to 10 feet due to the accuracy of the on-board GPS.  This keeps us from being able to overlay this dataset with other imagery; however, this will not affect our semantic segmentation results since the images have the same GSD and are registered relative to one another.  Our lab has recently acquired a higher-accuracy GPS and inertial navigation system (INS) to make overlaying multiple sensors (e.g., LIDAR, multispectral, and hyperspectral) possible for future collections.  In addition, the ability to quickly and accurately register multi-modal remote sensing data could enable the construction of larger DIRSIG scenes with improved spatial and spectral resolution.  The increase in quantity and quality of synthetic imagery could improve the network initialization scheme proposed in this paper for any sensor modality. 

\section{Conclusion}

We have shown that synthetic imagery can be used to assist the training of end-to-end semantic segmentation frameworks when there there is not enough annotated image data.  Our network initialization scheme has been shown to increase semantic segmentation performance when compared to traditional classifiers and unsupervised feature extraction techniques.  The features learned from the synthetic data successfully transfered to real-world imagery and prevented our RefineNet-Sim model from overfitting during training.  This work will enable remote sensing researchers to take advantage of advancements in deep-learning that were not previously available to them due to the lack of annotated image data.

In addition, we have introduced the RIT-18 dataset as an improved and more challenging benchmark for the semantic segmentation of MSI.  We will make this data available on the IEEE GRSS evaluation server in order to standardize the evaluation of new semantic segmentation frameworks. Although not the largest, RIT-18 is more practical because UAS platforms are easier/cheaper to fly.  RIT-18 is difficult to perform well on because of the high-spatial variability and unbalanced class distribution.

In the future, we hope to improve our model by 1) exploring deeper ResNet models; 2) using newer state-of-the-art convolution (ResNeXt~\cite{DBLP:journals/corr/XieGDTH16}) and segmentation models; 3) improving the inherent GSD of the DIRSIG scene; and 4) including additional diverse classes to the synthetic data.  The DCNN models we explored still produce classification maps with some salt-and-pepper label noise. This could be remedied in future work by newer end-to-end DCNN segmentation frameworks,  GEOBIA methods, or CRF-based algorithms. 
e These techniques should aid the development of more discriminative frameworks that yield superior performance.  

\section*{Acknowledgements}
We would like to thank Nina Raqueno, Paul Sponagle, Timothy Bausch, Michael McClelland II, and other members of the RIT Signature Interdisciplinary Research Area, UAS Research Laboratory that supported with data collection.  We would also like to thank Dr. Michael Gartley with DIRSIG support.

\newpage
\section*{Bibliography}
{\small
\bibliographystyle{elsarticle-num}
\bibliography{egbib.bib}
}

\newpage
\appendix

\section{RIT-18: Dataset Creation Details}
\setcounter{figure}{0}
\renewcommand\thefigure{\Alph{section}.\arabic{figure}}
\label{section:ortho}

\begin{figure}[h]
  \centering
  \includegraphics[width=0.95\linewidth]{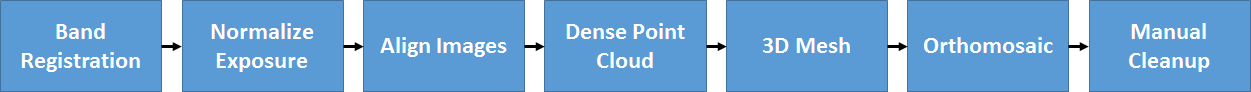}
  \caption{Orthomosaic Processing Pipeline \label{fig:pipeline}}
\end{figure}

In 2016, the Chester F. Carlson Center for Imaging Science established a new UAS laboratory to collect remote sensing data for research purposes.  This laboratory is equipped with several UAS payloads including RGB cameras,  MSI/HSI sensors, thermal imaging systems, and light detection and ranging (LIDAR).  This section will provide detail on how the RIT-18 dataset was created.  Fig. \ref{fig:pipeline} provides a macro-level view of our orthomosaic generation pipeline.

\begin{figure}[ht]
  \centering
  \subfigure[Affine ]{%
  \includegraphics[width=0.445\linewidth]{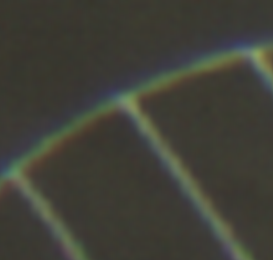}
  \label{fig:affinecal}}
  \subfigure[Perspective]{%
  \includegraphics[width=0.4655\linewidth]{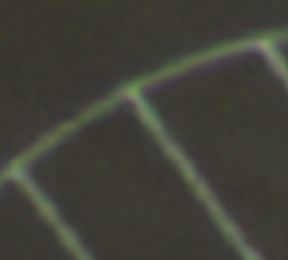}
  \label{fig:perspectivecal}}
  \caption{Difference between manufacturer's affine transformation and our perspective transformation.  The registration error in the affine transformation looks like a blue and red streak along the top and bottom of the parking lines, respectively.   \label{fig:transformation}
}
\end{figure}

The first step is to \textbf{co-register the images} from all six spectral bands.  For each collection campaign, we filtered out data not collected along our desired flight path (\emph{i.e.} takeoff and landing legs).  The six spectral images come from independent imaging systems, so they need to be registered to one another.  The manufacturer provided an affine transformation matrix that was not designed to work at the flying height this data was collected at, which caused noticeable registration error.  We used one of the parking lot images to develop a global perspective transformation for the other images in our dataset.  Fig. \ref{fig:transformation} illustrates the registration error caused by the affine transformation provided by the manufacturer and how this error can be reduced with our perspective transformation.

The global transformation worked well for some of the images, but there were registration errors in other parts of the scene, indicating that the transformation needs to be performed on a per-image basis.  This is critical when the wind causes excessive platform motion or when the UAV flies over trees.  This was done by 1) extracting SIFT features from each image, 2) using k-Nearest Neighbor to find the best matches, 3) keep the best matches, 4) use RANSAC to find the best homography, and 5) transforming the images with this homography using nearest-neighbor interpolation.  If there were no good matches, then we used the global perspective transformation.  This was done by matching SIFT features from each band to build custom homographies.  If a good homography could not be found, the global transformation was used instead.  This was common for homogenous scene elements, such as water, or repeating patterns, such as an empty parking lots.

The second step was to \textbf{normalize the exposure} for each band.  Each collected frame was acquired with a unique integration time (\emph{i.e.} auto exposure) and each band of the Tetracam Micro-MCA6 uses a different integration time proportional to the sensor's relative spectral response.  Another issue is that each image, including each band, is collected with a different integration time.  The longer integration times required for darker images, especially over water scenes, resulted in blur caused by platform motion.  We normalized each image with its corresponding integration time and then contrast-stretched the image back to a 16-bit integer using the global min/max of the entire dataset.  The original images are 10-bit, but the large variation in integration time groups most of the data to lower intensity ranges.  We extended the dynamic range of the orthomosaic by stretching the possible quantized intensity states.  We generated the othomosaics mostly because of time-constraints; however, we are also interested in evaluating models that train on small quantities of annotated imagery.

The remaining steps were performed using Agisoft PhotoScan~\cite{photoscan2016}.  Photoscan uses the 720nm band to perform most of the procedures, and then the remaining spectral channels are brought back in the orthomosaic generation step.  The PhotoScan workflow involves:

\begin{enumerate}
\setlength\itemsep{0em}
\item \textbf{Image Alignment:} Find key points in the images and match them together as tie-points.
\item \textbf{Dense Point Cloud Reconstruction:} Use structure from motion (SfM) to build a dense point cloud from the image data.
\item \textbf{3D Mesh:} Build a 3D mesh and corresponding UV texture map from the dense point cloud.
\item \textbf{Orthomosaic:} Generate an orthomosaic onto the WGS-84 coordinate system using the mesh and image data.
\item \textbf{Manual Clean-up:} Manually correct troublesome areas by removing photographs caused by motion blur or moving objects.
\end{enumerate}

PhotoScan can generate high-quality orthomosaics, but manual steps were taken to ensure the best quality.  First, not all of the images were in focus; and although PhotoScan has an image quality algorithm, we opted to manually scan and remove the defocused images.  Second, the 3D model that the orthomosaic is projected onto is built from structure-from-motion.  Large objects that move over time, such as tree branches blowing in the wind, or vehicles moving throughout the scene, will cause noticeable errors.  This is corrected by highlighting the affected region and manually selecting a single (or a few) alternative images that will be used to generate that part of the orthomosaic, as opposed to those automatically selected.  

\section{RIT-18: Class Descriptions}
\label{section:classes}

\subsection{Water/Beach Area}
The two classes for water are lake and pond.  The lake class is for Lake Ontario, which is north of the beach.  The pond water class is for the small inland pond, present in all three folds, which is surrounded by marsh and trees.  Along Lake Ontario is a sand/beach class.  This class also includes any spot where sand blew up along the asphalt walking paths.  Along the beach are some white-painted, wooden lifeguard chairs.  The buoy class is for the water buoys present in the water and on the beach.  They are very small, primarily red and/or white, and assume various shapes.  The rocks class is for the large breakwater along the beach.

\subsection{Vegetation}
There are three vegetation classes including grass, trees, and low-level vegetation.  The tree class includes a variety of trees present in the scene.  The grass includes all pixels on the lawn.  There are some mixtures present in the grass (such as sand, dirt, or various weeds), so the classification algorithm will need to take neighboring pixel information into account.  The grass spots on the beach and asphalt were labeled automatically using a normalized-difference vegetation index (NDVI) metric.  Grass spots that were missed were manually added. The low-level vegetation class includes any other vegetation, including manicured plants, around the building or the marsh next to the pond. 

\subsection{Roadway}
The asphalt class includes all parking lots, roads, and walk-ways made from asphalt; but the cement and stone paths around the buildings are not consistent between different folds, so they remain labeled as the background class.  The road marking class is for any painted asphalt surface including parking/road lanes.  This class was automatically labeled with posteriori, but there were a few parking lines in the shade that needed to be manually added.  The road markings in the validation image are sharper than those depicted in the training image since the park repainted the lines between collects.  The vehicle label includes any car, truck, or bus.

\subsection{Underrepresented Classes}
Underrepresented classes, which may be small and/or appear infrequently, will be difficult to identify.  Since some of the land cover classes are massive in comparison, the mean-class accuracy metric will be the most important during the classification experiments in Table \ref{table:results}.  Small object classes, such as person and picnic table, represent only a minute fraction of the image and should remain very difficult to correctly classify.  These small objects will be surrounded by larger classes and may even hide in the shade.

There are also a few classes that are only present in the scene a couple of times, such as the white/black wood targets, orange UAS landing pad, lifeguard chair, and buildings.  The building class is primarily roof/shingles of a few buildings found throughout the scene.  The similarity between the white wooden target and the lifeguard chair should make semantic information in the scene vital to classification accuracy.  There is only a single instance of the orange UAS landing pad in every fold.  The black and white targets are not present in the validation fold, which could make it difficult to cross-validate for a model that can correctly identify them.

\section{SCAE Architecture}
\setcounter{figure}{0}    
\label{section:scae}
SCAE, illustrated in Fig. \ref{fig:scae}, is another unsupervised spatial-spectral feature extractor.  SCAE has a deeper neural network architecture than MICA and is capable of extracting higher-level features \cite{Kemker16}.  The architecture used in this paper involves three individual convolutional autoencoders (CAEs) that are trained independently.  The input and output of the first CAE is a collection of random image patches from the training data, and the input/output of the subsequent CAEs are the features from the last hidden layer of the previous CAE.  The output of all three CAEs are concatenated in the feature domain, mean-pooled, and the dimensionality is reduced to 99\% of the original variance using WPCA.  The final feature response is scaled to zero-mean/unit-variance and then passed to a traditional classifier. Here, we use an MLP with one hidden layer. 

\begin{figure}[ht!]
\begin{center}
   \includegraphics[width=0.60\linewidth]{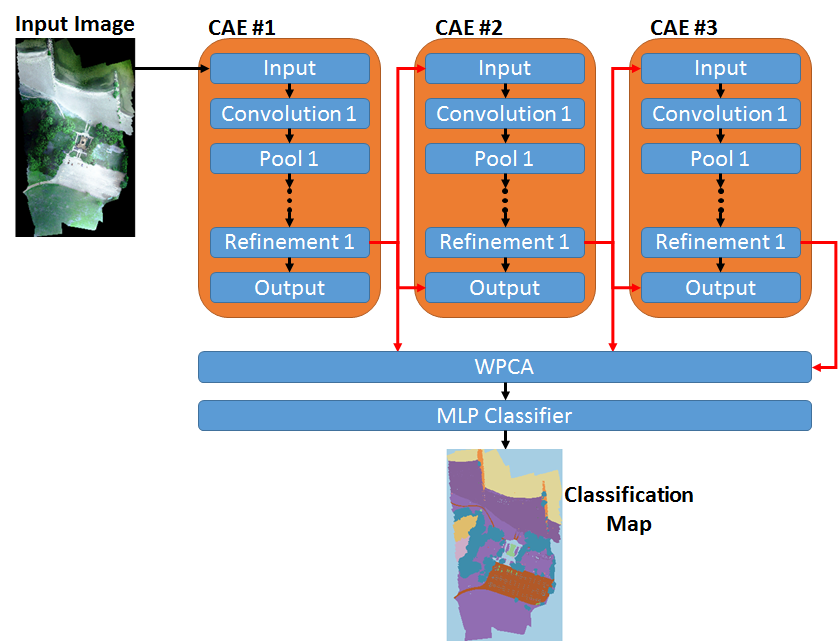}
\end{center}
   \caption{The SCAE model used in this paper.  Architecture details for each CAE are shown in Fig. \ref{fig:scae_parts}.}
\label{fig:scae}
\end{figure}

\begin{figure*}[t]
\begin{center}
   \includegraphics[width=0.60\linewidth]{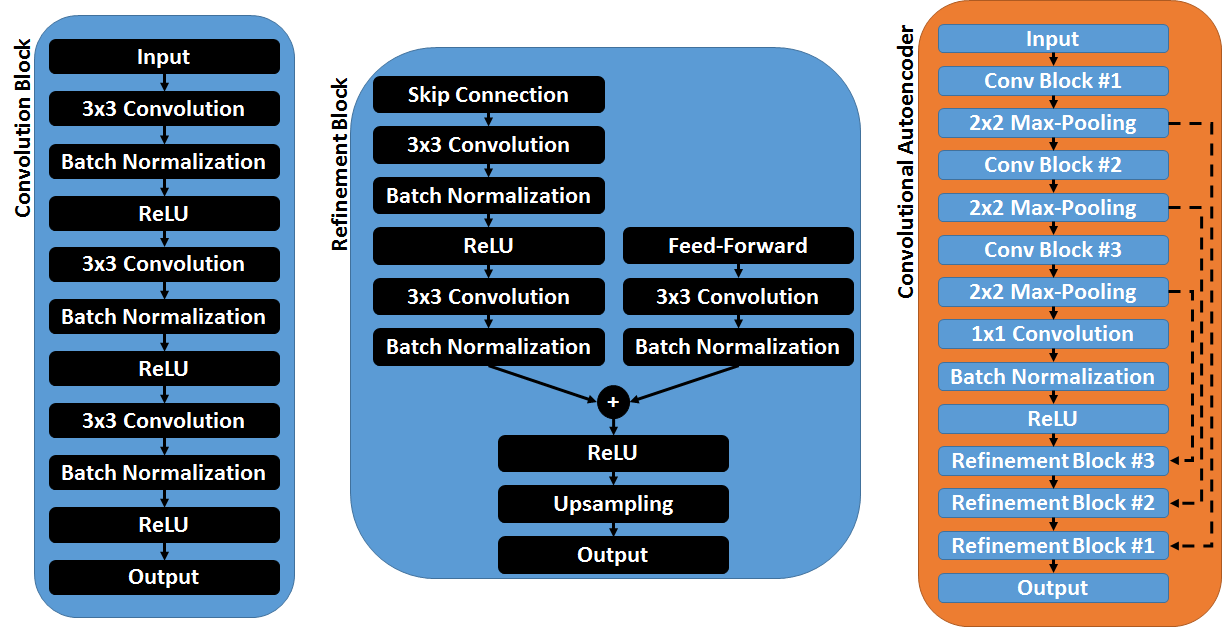}
\end{center}
   \caption{The convolutional autoencoder (CAE) architecture used in SCAE.  This CAE is made up of several convolution and refinement blocks.  The SCAE model in this paper uses three CAEs.}
\label{fig:scae_parts}
\end{figure*}

The architecture of each CAE is illustrated in Fig. \ref{fig:scae_parts}.  Each CAE contains a small feed-forward network consisting of multiple convolution and max-pooling operations. The feature response is reconstructed with symmetric convolution and upsampling operations.  The reconstruction error is reduced by using skip connections from the feed-forward network, inspired by \cite{PinheiroLCD16}.

The SCAE model used in this paper has the same architecture shown in Fig. \ref{fig:scae} and \ref{fig:scae_parts}.  It was trained with 30,000 $128\times 128$ image patches randomly extracted from the training and validation datasets.  Each CAE was trained individually with a batch size of 128.  There are 32 units in the first convolution block, 64 units in the second convolution block, 128 units in the third, and 256 units in the $1 \times 1$ convolution.  The refinement blocks have the same number of units as their corresponding convolution block, so the last hidden layer has 32 features.  

After training, the whole image is passed through the SCAE network to generate three $N\times 32$ feature responses.  These feature responses are concatenated, convolved with a $5 \times 5$ mean-pooling filter, and then reduced to 99\% of the original variance using WPCA.  The final feature response is passed to a MLP classifier with the same architecture used by the MICA model.

\section{Additional Results}
\label{appendix:results}

\begin{figure}[ht]
  \centering
  \subfigure[Sharpmask-Rdm ]{%
  \includegraphics[width=0.445\linewidth]{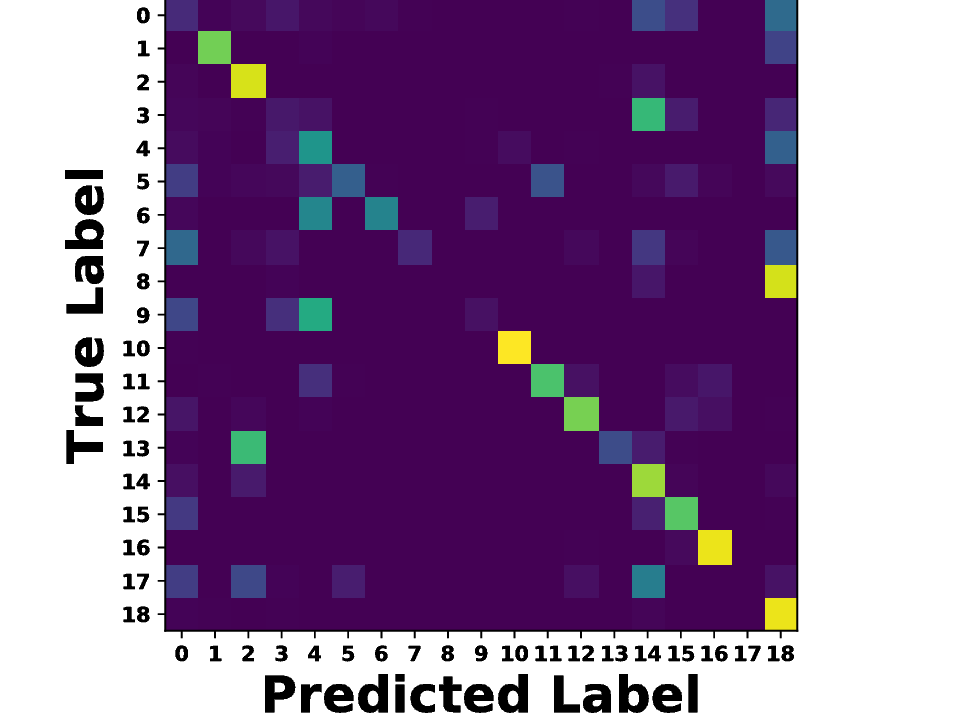}
  \label{fig:sharpmask_cm_nodirsig}}
  \subfigure[Sharpmask-Sim]{%
  \includegraphics[width=0.4655\linewidth]{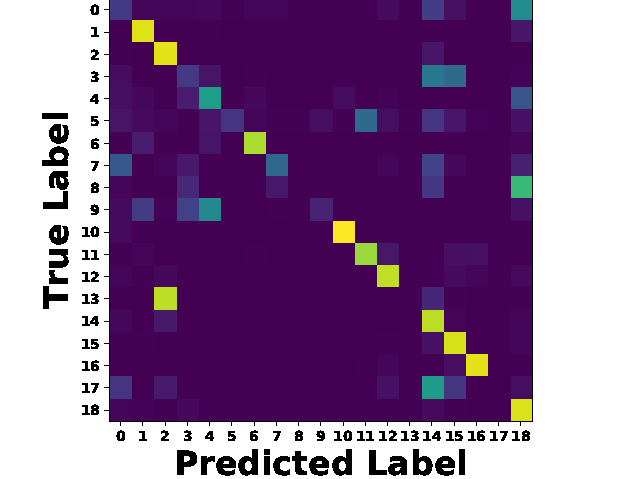}
  \label{fig:sharpmask_cm_dirsig}}
   \subfigure[RefineNet-Rdm ]{%
  \includegraphics[width=0.445\linewidth]{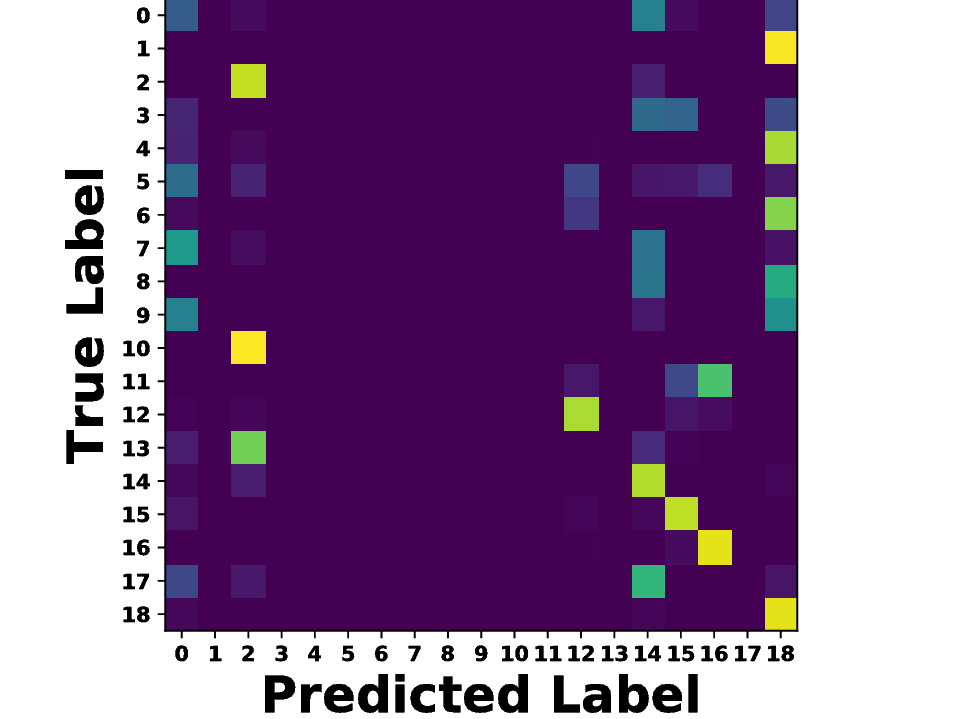}
  \label{fig:refinenet_cm_nodirsig}}
  \subfigure[RefineNet-Sim]{%
  \includegraphics[width=0.4655\linewidth]{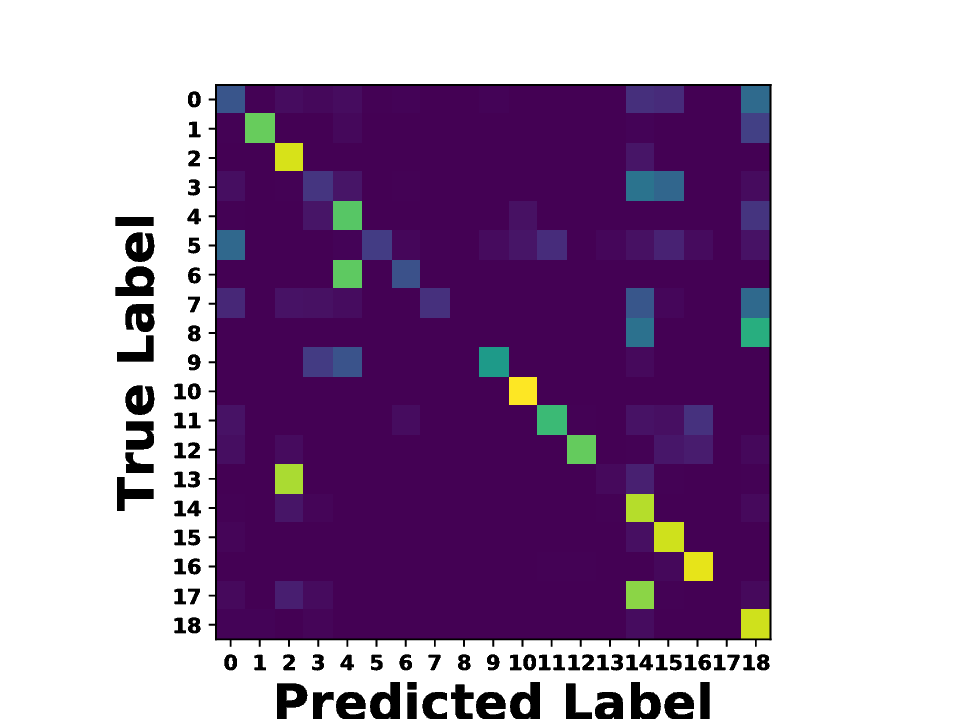}
  \label{fig:refinenet_cm_dirsig}}
  
  \caption{Heatmap visualization of the confusion matrices for all four models.  Each row is normalized to itself to highlight the most common errors.\label{fig:cm}
}
\end{figure}

Fig. \ref{fig:cm} shows a heat-map of the confusion matrices for SharpMask and RefineNet - with and without DIRSIG pre-training.  These results were generated from the prediction map on the RIT-18 test set.  Each row is normalized to show the most common prediction for each class.  The brighter (yellow) squares indicate a high classification accuracy for that particular class.  A strong (bright) diagonal for each confusion matrix shows that the model is doing well, and strong off-diagonal elements indicate that where the model is commonly misclassifying a particular class.  The RefineNet-Rdm model (Fig. D.\ref{fig:refinenet_cm_nodirsig}) is clearly overfitting to the classes with more training samples; but when the model is pre-trained with DIRSIG data (Fig. D.\ref{fig:refinenet_cm_dirsig}; it does a better job classifying the test set.

\begin{figure}[ht]
  \centering
  \subfigure[Loss]{%
  \includegraphics[width=0.9\linewidth]{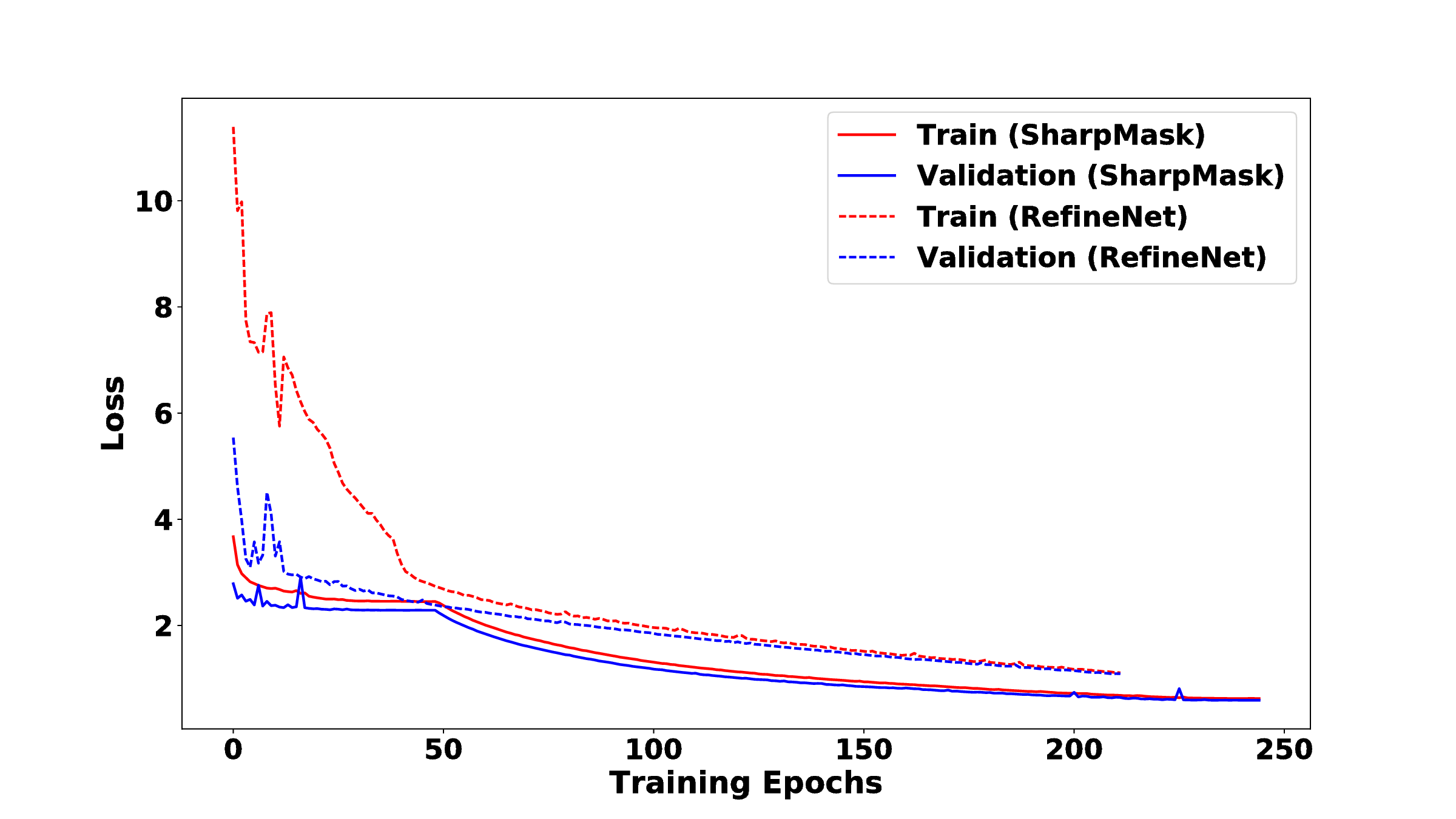}
  \label{fig:loss}}
  \subfigure[Overall Accuracy]{%
  \includegraphics[width=0.9\linewidth]{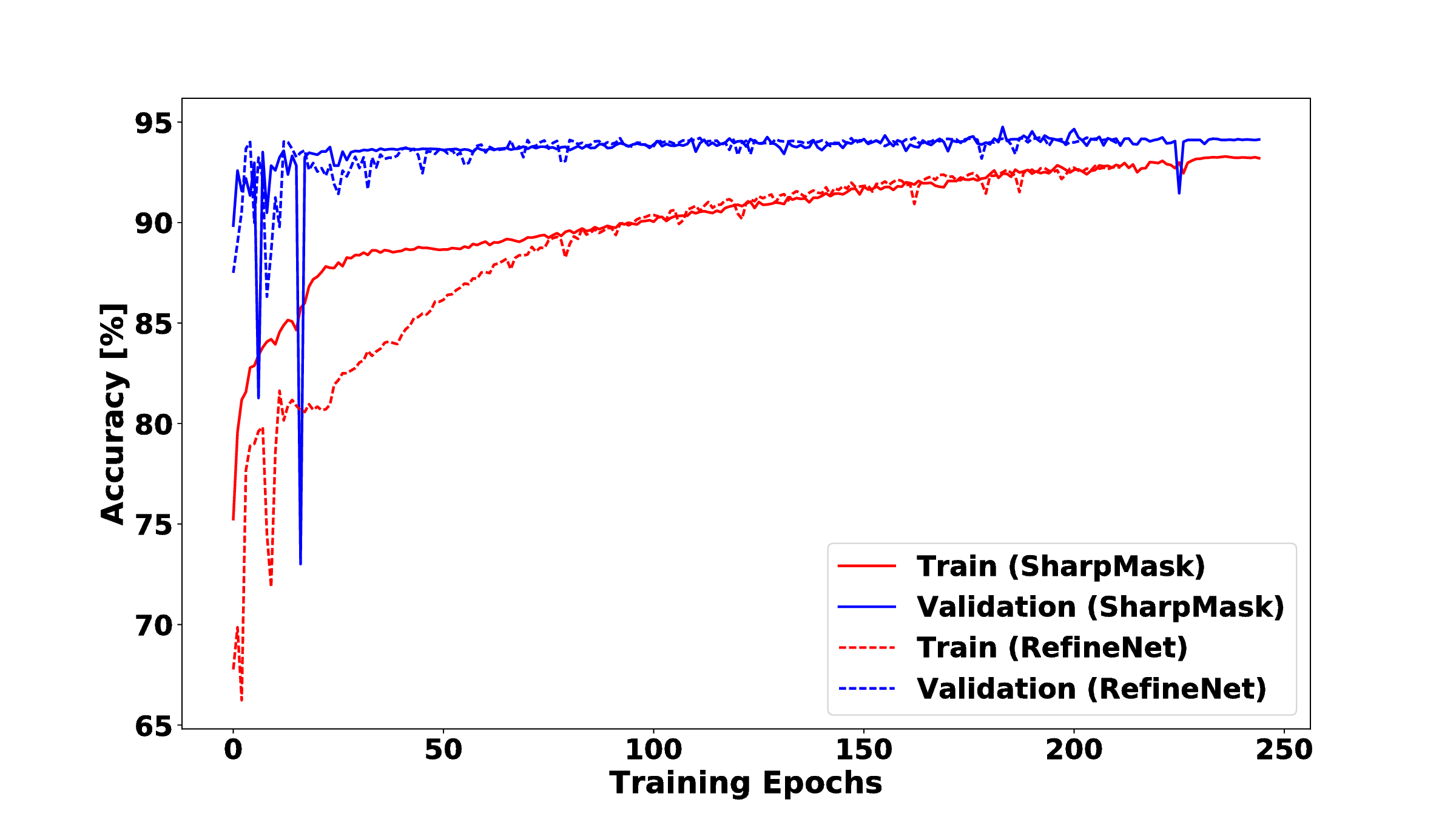}
  \label{fig:acc}}
  \caption{Training (red) and validation (blue) loss and accuracy plots for SharpMask (solid line) and RefineNet (dashed line) models with DIRSIG weight initialization.\label{fig:loss_acc}
}
\end{figure}

Fig. \ref{fig:loss_acc} shows the loss and accuracy curves for SharpMask-Sim (solid) and RefineNet-Sim (dashed).  Fig. D.\ref{fig:loss} shows that the training and validation loss stay very close, which indicates that the model is not overfitting to the training data.   

\begin{figure}[ht]
  \centering
  \includegraphics[width=0.9\linewidth]{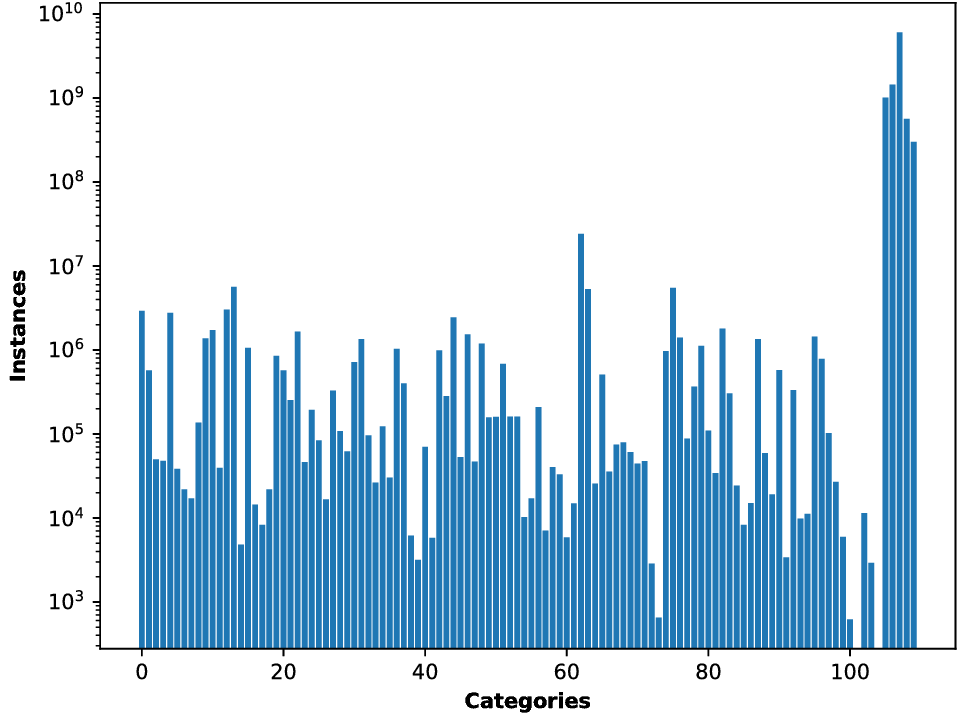}
  \caption{Histogram of class distribution for DIRSIG training set.\label{fig:dirsig_stats}
}
\end{figure}

Fig. \ref{fig:dirsig_stats} shows the class distribution of the generated DIRSIG dataset.  This dataset is also unbalanced, which is why we use a weighted loss function (Equation \ref{eq:sample_weights}) to pre-train the ResNet-50 DCNN.

\end{document}